%% file: SSTC.tex
\documentclass[lettersize,journal]{IEEEtran}
\usepackage{amsmath,amsfonts}
\usepackage{algorithmic}
\usepackage{algorithm}
\usepackage{array}
\usepackage[caption=false,font=normalsize,labelfont=sf,textfont=sf]{subfig}
\usepackage{textcomp}
\usepackage{stfloats}
\usepackage{url}
\usepackage{verbatim}
\usepackage{graphicx}
\usepackage{cite}
\usepackage{xcolor}
\usepackage{times}
\usepackage{soul}
\usepackage{url}
\usepackage{comment}
\usepackage{mathtools}

\usepackage{booktabs} 
\usepackage[switch]{lineno}

\usepackage{color}
\usepackage{soul}
\usepackage[utf8]{inputenc}
\usepackage{hyperref}

\usepackage{multirow}
\usepackage{makecell}
\usepackage{amsmath,amssymb}
\usepackage{mathrsfs}
\usepackage{latexsym}
\usepackage{varioref}
\usepackage{xspace}
\usepackage{bm}
\usepackage{enumitem}
\usepackage{pifont}
\usepackage{amsthm}
\usepackage{cleveref}
\usepackage[figuresright]{rotating}

\newcommand{\eg}{\emph{e.g., }}
\newcommand{\ie}{\emph{i.e., }}

\newcommand{\baby}{S$^2$\textsc{tc-bdd}\xspace}
\newcommand{\babys}{S$^2$\textsc{tc-bdd-s}\xspace}
\newcommand{\babyf}{S$^2$\textsc{tc-bdd-f}\xspace}

\newcommand{\norm}[1]{\lVert#1\rVert}

\newcommand{\tabincell}[2]{\begin{tabular}{@{}#1@{}}#2\end{tabular}}

\theoremstyle{plain}

\theoremstyle{definition}

\begin{document}

\title{Semi-Supervised Learning with Balanced Deep Representation Distributions}

\author{Changchun Li, Ximing Li$^*$, Bingjie Zhang, Wenting Wang, and Jihong Ouyang

\thanks{Manuscript received April 19, 2021; revised August 16, 2021. This work was supported in part by the National Natural Science Foundation of China under Grant 62276113; in part by the China Postdoctoral Science Foundation under Grant 2022M721321; and in part by the Energy Administration of Jilin Province under Grant 3D516L921421.}
\thanks{Changchun Li, Ximing Li \emph{(corresponding author, denoted by $*$)}, Bingjie Zhang and Jihong Ouyang are with College of Computer Science and Technology, Jilin University, Changchun 130012, China, also with the Key Laboratory of Symbolic Computation and Knowledge Engineering of Ministry of Education, Jilin University, Changchun 130012, China (e-mail: changchunli93@gmail.com; liximing86@gmail.com; zhangbj21@mails.jlu.edu.cn; ouyj@jlu.edu.cn).}
\thanks{Wenting Wang is with the Department of Computer Science, University of Texas at Dallas, Richardson, TX, USA (e-mail: Wenting.Wang@UTDallas.edu).}
}

\markboth{Journal of \LaTeX\ Class Files,~Vol.~14, No.~8, August~2021}%
{Shell \MakeLowercase{\textit{et al.}}: Semi-Supervised Learning with Balanced Deep Representation Distributions}


\maketitle

\begin{abstract}
\input{Sec_Abstract}
\end{abstract}

\begin{IEEEkeywords}
Semi-supervised learning, text classification, balanced deep representation distributions, self-training, multi-label learning
\end{IEEEkeywords}

\input{Sec_Introduction}

\input{Sec_RelatedWork}
\input{Sec_Method}
\input{Sec_Experiment}
\input{Sec_Discussion}
\input{Sec_Conclusion}

\bibliographystyle{IEEEtran}
\bibliography{SSTC}

\vfill

\end{document}

%% file: Sec_Abstract.tex
Semi-Supervised Text Classification (SSTC) mainly works under the spirit of self-training. They initialize the deep classifier by training over labeled texts; and then alternatively predict unlabeled texts as their pseudo-labels and train the deep classifier over the mixture of labeled and pseudo-labeled texts. Naturally, their performance is largely affected by the accuracy of pseudo-labels for unlabeled texts. Unfortunately, they often suffer from low accuracy because of the \emph{margin bias} problem caused by the large difference between representation distributions of labels in SSTC. To alleviate this problem, we apply the angular margin loss, and perform several Gaussian linear transformations to achieve balanced label angle variances, \ie the variance of label angles of texts within the same label. More accuracy of predicted pseudo-labels can be achieved by constraining all label angle variances balanced, where they are estimated over both labeled and pseudo-labeled texts during self-training loops. With this insight, we propose a novel SSTC method, namely Semi-Supervised Text Classification with Balanced Deep representation Distributions (\baby). 
We implement both multi-class classification and multi-label classification versions of \baby by introducing some pseudo-labeling tricks and regularization terms.
To evaluate \baby, we compare it against the state-of-the-art SSTC methods. Empirical results demonstrate the effectiveness of \baby, especially when the labeled texts are scarce.

%% file: Sec_Introduction.tex
\section{Introduction}
\label{1}

\begin{figure}
\includegraphics[width=0.45\textwidth]{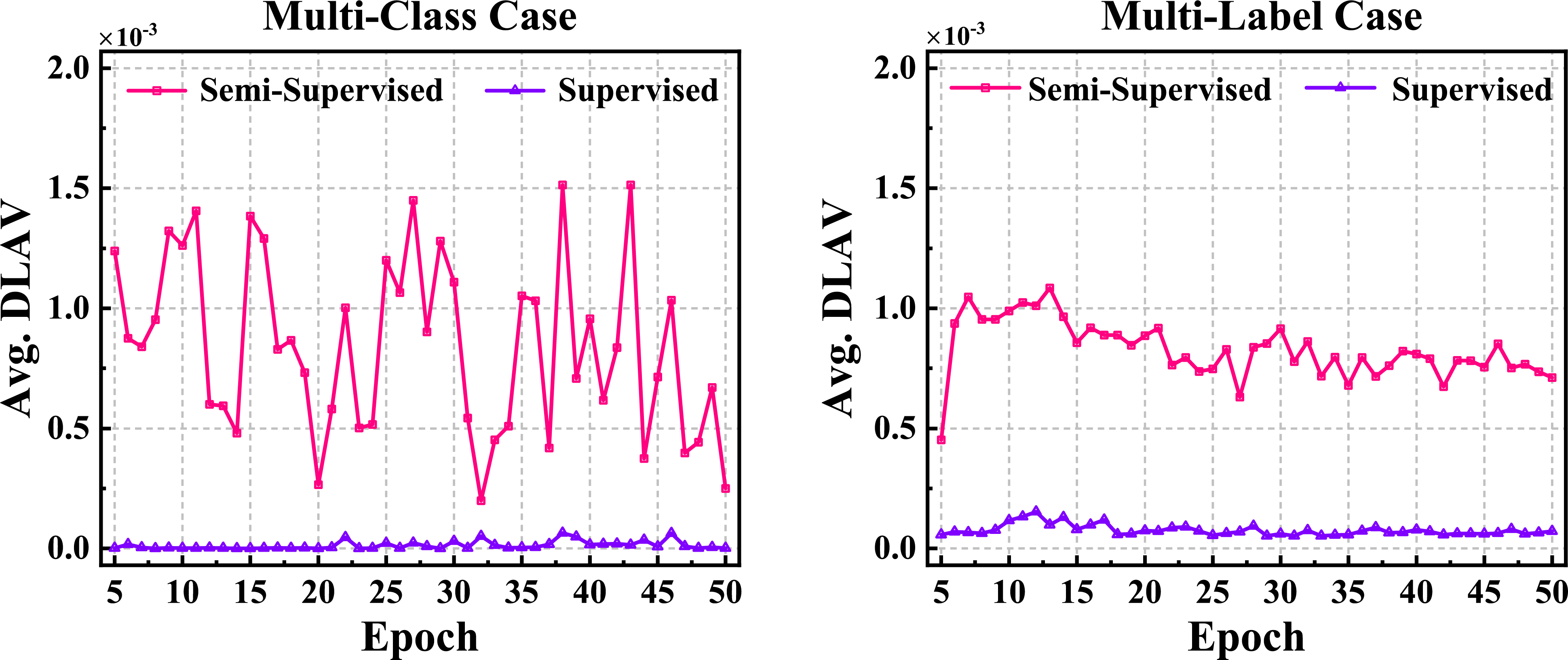}
\centering
\caption{The average difference of label angle variances (Avg.DLAV) computed in semi-supervised and supervised manners across \emph{AG News} (\textbf{Multi-Class Case}) and \emph{AAPD} (\textbf{Multi-Label Case}), respectively.}
\label{Fig1}
\end{figure}

\IEEEPARstart{S}{emi-\textbf{S}upervised} \textbf{L}earning (\textbf{SSL}) refers to the paradigm of learning with labeled as well as unlabeled data to perform certain applications \cite{SSL2020,TCYB5,TCYB6}, and has received great research interests in the past decades \cite{SSL2020,TCYB1,TCYB2,TCYB3,TCYB4,TCYB7,TCYB8,TCYB9}. Especially, developing effective SSL models for classifying text data has long been a goal for the studies of natural language processing, because labeled texts are difficult to collect in many real-world scenarios. Formally, this research topic is termed as \textbf{S}emi-\textbf{S}upervised \textbf{T}ext \textbf{C}lassification (\textbf{SSTC}), which nowadays draws much attention from the community \cite{CVT2018,VAM2019,MixText2020,SoftMatch2023}.  

To our knowledge, the most recent SSTC methods mainly borrow ideas from the successful patterns of supervised deep learning, such as pre-training and fine-tuning \cite{Dai2015,ULMFit2018,ELMO2018,VAM2019,BERT2019}. Generally, those methods perform deep representation learning on unlabeled texts followed by supervised learning on labeled texts. However, a drawback is that they separately learn from the labeled and unlabeled texts, where, specifically, the deep representations are trained without using the labeling information, resulting in potentially less discriminative representations as well as worse performance.

To avoid this problem, other SSTC methods combine the traditional spirit of self-training with deep learning, which jointly learns the deep representation and classifier using both labeled and unlabeled texts in a unified framework \cite{VAT2017,VAT2019,LSTMMIXED2019,UDA2020,MixText2020,FreeMatch2023,SoftMatch2023}. To be specific, this kind of method initializes a deep classifier, \eg BERT \cite{BERT2019} with Angular Margin (AM) loss \cite{CosFace2018}, by training over labeled texts only; and then it alternatively predicts unlabeled texts as their pseudo-labels and trains the deep classifier over the mixture of labeled and pseudo-labeled texts. Accordingly, both labeled and unlabeled texts can directly contribute to deep classifier training. 

\begin{figure}
\includegraphics[width=0.45\textwidth]{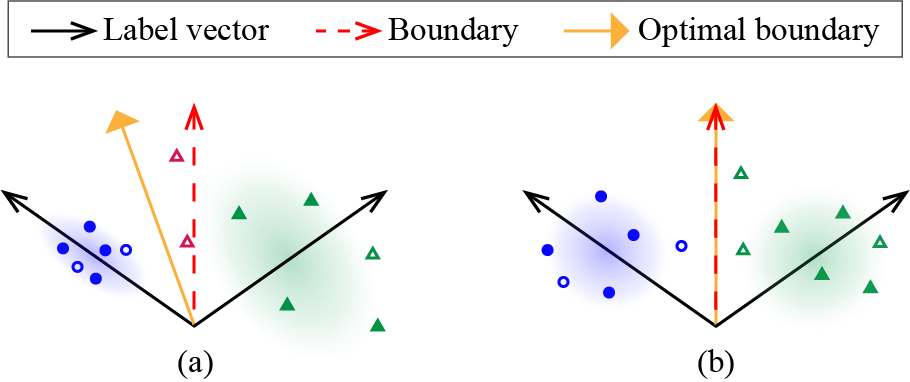}
\centering
\caption{Let solid circles and triangles denote labeled positive and negative texts, and hollow ones denote corresponding unlabeled texts. (a) The large difference between label angle variances results in the margin bias. Many unlabeled texts (in red) can be misclassified. (b) Balancing the label angle variances can eliminate the margin bias. Best viewed in color.}
\label{example}
\end{figure}

Generally speaking, for deep self-training methods, one significant factor of performance is the accuracy of pseudo-labels for unlabeled texts. Unfortunately, they often suffer from low accuracy, and one major reason is the \textbf{\emph{margin bias}} problem caused by \emph{the large difference between representation distributions of labels in SSTC.} To interpret this problem, we look around the AM loss with respect to the \textbf{label angle}, \ie the angles between deep representations of texts and weight vectors of labels. For unlabeled texts, the pseudo-labels are predicted by only ranking the label angles, but neglecting the difference between \textbf{label angle variances}, \ie the variance of label angles of texts within the same label, which might be much larger in SSL as illustrated in Fig.\ref{Fig1}. In this context, the boundary of AM loss is actually not the optimal one, potentially resulting in lower accuracy for pseudo-labels, as illustrated in Fig.\ref{example}(a).

To alleviate the aforementioned problem, we propose a novel SSTC method built on BERT with AM loss, namely \textbf{S}emi-\textbf{S}upervised \textbf{T}ext \textbf{C}lassification with \textbf{B}alanced \textbf{D}eep representation \textbf{D}istributions (\textbf{\baby}). Most specifically, in \baby, we suppose that the label angles are drawn from each label-specific Gaussian distribution. Therefore, for each text, we can apply linear transformation operations to balance the label angle variances, \ie the variance of label angles of texts within the same label. This is equivalent to moving the boundary to the optimal one, so as to eliminate the margin bias (see examples in Fig.\ref{example}(b)). Upon this idea, we design a novel BDD loss by incorporating the Gaussian linear transformations into AM loss and construct the novel SSTC framework \baby based on the designed BDD loss and deep self-training spirit, where each label angle variance can be estimated over both labeled and pseudo-labeled texts during the self-training loops. We implement both multi-class classification and multi-label classification versions of \baby by introducing some pseudo-labeling tricks and regularization terms. We evaluate the proposed \baby method by comparing the most recent deep SSTC methods. Experimental results indicate the superior performance of \baby even with very few labeled texts on both multi-class classification and multi-label classification scenarios.

To sum up, the major contributions of this paper are presented as follows:
\begin{itemize}
    
    \item We develop a novel SSTC method, namely \baby, by constraining deep representation distributions balanced, based on the identified margin bias problem caused by the large difference between representation distributions of labels in SSTC.

    \item We design a novel BDD loss by incorporating several Gaussian linear transformations into AM loss to achieve balanced label angle variances, \ie the variance of label angles of texts within the same label, so as to eliminate the margin bias.

    \item We implement both multi-class classification and multi-label classification versions of \baby by introducing some pseudo-labeling tricks and regularization terms.

    \item We conduct extensive experiments to evaluate \baby, and the results demonstrate the superior performance of \baby even with very few labeled texts.

\end{itemize}

\vspace{5pt}
\noindent\textbf{Remark.} 
This paper is an extension version of our previous conference work \cite{BDD2021}.
The extended works include (1) a novel multi-class classification version of \baby; (2) a novel multi-label classification version of \baby; (3) a novel alternative optimization procedure for the multi-label classification version of \baby; and (4) more experiments to evaluate the effectiveness of \baby.

%% file: Sec_RelatedWork.tex
\section{Related Work}
\label{2}


\subsection{Semi-Supervised Multi-Class Text Classification}
The pre-training and fine-tuning framework has lately shown impressive effectiveness on a variety of tasks \cite{Dai2015,GPT2018,BERT2019,XLNet2019,GPT22019,GPT32020,MixText2020}.
They mainly perform deep representation learning on generic data, followed by supervised learning for downstream tasks.
Several semi-supervised multi-class text classification methods are built on this framework \cite{Dai2015,ULMFit2018,ELMO2018,VAM2019,BERT2019}. For instance, the VAriational Methods for Pretraining In Resource-limited Environments (VAMPIRE) \cite{VAM2019} first pre-trains a Variational Auto-Encoder (VAE) model on unlabeled texts, and then trains a classifier on the augmentation representations of labeled texts computed by the pre-trained VAE. However, the VAE model is trained without using the labeling information, resulting in potentially less discriminative representations for labeled texts.

Recent works on semi-supervised multi-class text classification mainly focus on deep self-training \cite{LSTMMIXED2019,UDA2020,SALNet2021,PCSM2022,PGPL2023}, which can jointly learn deep representation and classifier using both labeled and unlabeled texts in a unified framework. It is implemented by performing an alternative process, in which the pseudo-labels of unlabeled texts are updated by the current deep classifier, and then the deep classifier is retrained over both labeled and pseudo-labeled texts. There exist two major practices:
One line of the methods \cite{VAT2017,VAT2019,LSTMMIXED2019} follows the philosophy of making the classifier robust against random and local perturbation. 
It first generates the predictions of original texts with the current deep classifier and then trains the deep classifier by utilizing a consistency loss between the original predictions and the outputs of deep classifier over noise texts by applying local perturbations to the embeddings of original texts. 
Another line of the methods such as Unsupervised Data Augmentation (UDA) \cite{UDA2020} and FixMatch \cite{FixMatch2020}, rather than applying local perturbations, employ weak and strong data augmentation techniques such as back translations, and exploit consistency loss between the predictions of weakly- and strongly-augmented versions of unlabeled texts with high confidences. 
Some works \cite{FreeMatch2023,SoftMatch2023} further propose self-adaptive approaches to exploit unlabeled texts more effectively, respectively.
Otherwise, the work \cite{CVT2018} exploits cross-view training by matching the predictions of auxiliary prediction modules over the restricted views of unlabeled texts (\eg only part of the sentence) with ones of primary prediction modules over the corresponding full views.
The current MetaExpert \cite{MetaExpert2025} focuses on the long-tailed semi-supervised learning with distribution mismatch, and dynamically assigns a suitable expert (classifier) to each text based on the class membership estimated by an expert assignment strategy, to produce high-quality pseudo-labels during the training procedure and predictions in testing.


\subsection{Semi-Supervised Multi-Label Text Classification}
Recently deep learning methods have achieved remarkable success in multi-label text classification. The most straightforward strategy \cite{CNNSC2014,RNNTC2016,BILLC2016,DLEMLTC2017,SGM2018,DEMLL2018,TPT2020,LightXML2021,FMRTF2021} is to extract deep representations of texts with deep neural networks, such as CNN, RNN, and transformer-based models, and then induce classification models with traditional multi-label learning approaches. Some other methods \cite{AttentionXML2019,LSDR2019,LSDGNN2021,BSF2022,CLLS2022} focus on learning label-specific representations to facilitate the discrimination of each label. 
The last group of methods designs specific architectures \cite{DVAE2020,CGP2021,HOTVAE2021} or training strategies \cite{GMVACL2022,CLENNM2022} to exploit the label corrections. Variational Autoencoder based models \cite{DVAE2020,HOTVAE2021} and contrastive learning based strategies \cite{GMVACL2022,CLENNM2022} are two groups of representative methods. Besides, some sophisticated loss functions \cite{BMMLTC2021,ALMLL2021,TWMLL2023} are also designed to improve the performance of multi-label learning.

To the best of our knowledge, relatively few works \cite{DRML2020,DRML2021,SSMLLDSGM2020,SSMLLGD2021,CAP2023} in the recent literature have been proposed to exploit the performance of deep models in semi-supervised multi-label learning scenarios. The Semi-supervised Dual Relation Learning (SDRL) \cite{DRML2020,DRML2021} employs a two-classifier structure with a label relation network to exploit feature-level and label-level correlations. It first performs a two-classifier domain adaption procedure to mitigate the feature distribution shift between labeled and unlabeled instances, and then trains the classifiers and label relation network with both labeled and pseudo-labeled instances, which are selected according to the prediction difference of two classifiers. The method proposed in \cite{SSMLLDSGM2020} employs a deep sequential generative model to address the noisy labels collected by crowdsourcing and unlabeled data simultaneously. The work in \cite{SSMLLGD2021} leverages graph neural networks with label embeddings, and designs joint variational representation and confidence-rated margin ranking modules to handle semi-supervised multi-label data with graph structures. The most related work is Class-Aware Pseudo-Labeling (CAP) method \cite{CAP2023}. It performs pseudo-labeling in a class-aware manner and employs the deep self-training strategy with the asymmetric loss \cite{ALMLL2021} for semi-supervised multi-label learning.

Orthogonal to the aforementioned self-training semi-supervised multi-class text classification and semi-supervised multi-label text classification methods, our \baby further identifies the margin bias problem caused by the large difference between representation distributions in SSTC, and designs a novel BDD loss with several Gaussian linear transformations to constrain label representation distributions to be balanced. This is beneficial for more accurate pseudo-labels for unlabeled texts, so as to boost the performance of both semi-supervised multi-class text classification and semi-supervised multi-label text classification tasks.


%% file: Sec_Method.tex
\section{The Proposed \baby Method}
\label{3}

In this section, we describe the proposed deep self-training SSTC method, namely \textbf{S}emi-\textbf{S}upervised \textbf{T}ext \textbf{C}lassification with \textbf{B}alanced \textbf{D}eep representation \textbf{D}istributions (\textbf{\baby}). 

\vspace{5pt}
\noindent\textbf{Formulation of SSTC.}
Consider a training dataset $\mathcal{D}$ consisting of a limited labeled text set $\mathcal{D}_{l}=\{(\mathbf{x}_i^l,\mathbf{y}_i^l)\}_{i=1}^{i=N_l}$ and a large unlabeled text set $\mathcal{D}_{u}=\{\mathbf{x}_j^u\}_{j=1}^{j=N_u}$. Specifically, let $\mathbf{x}_i^l$ and $\mathbf{x}_j^u$ denote the word sequences of labeled and unlabeled texts, respectively; and $\mathbf{y}_i^l\in\{0,1\}^K$ the corresponding label vector of $\mathbf{x}_i^l$, where $y_{ik}^l=1$ if the text is associated with the $k$-th label, or $y_{ik}^l=0$ otherwise. We declare that $N_l$, $N_u$, and $K$ denote the numbers of labeled texts, unlabeled texts and category labels, respectively. In this paper, we focus on the paradigm of inductive SSTC, whose goal is to learn a classifier from the training dataset $\mathcal{D}$ with both labeled and unlabeled texts. 


\begin{figure*}[t]
  \includegraphics[width=0.95\textwidth]{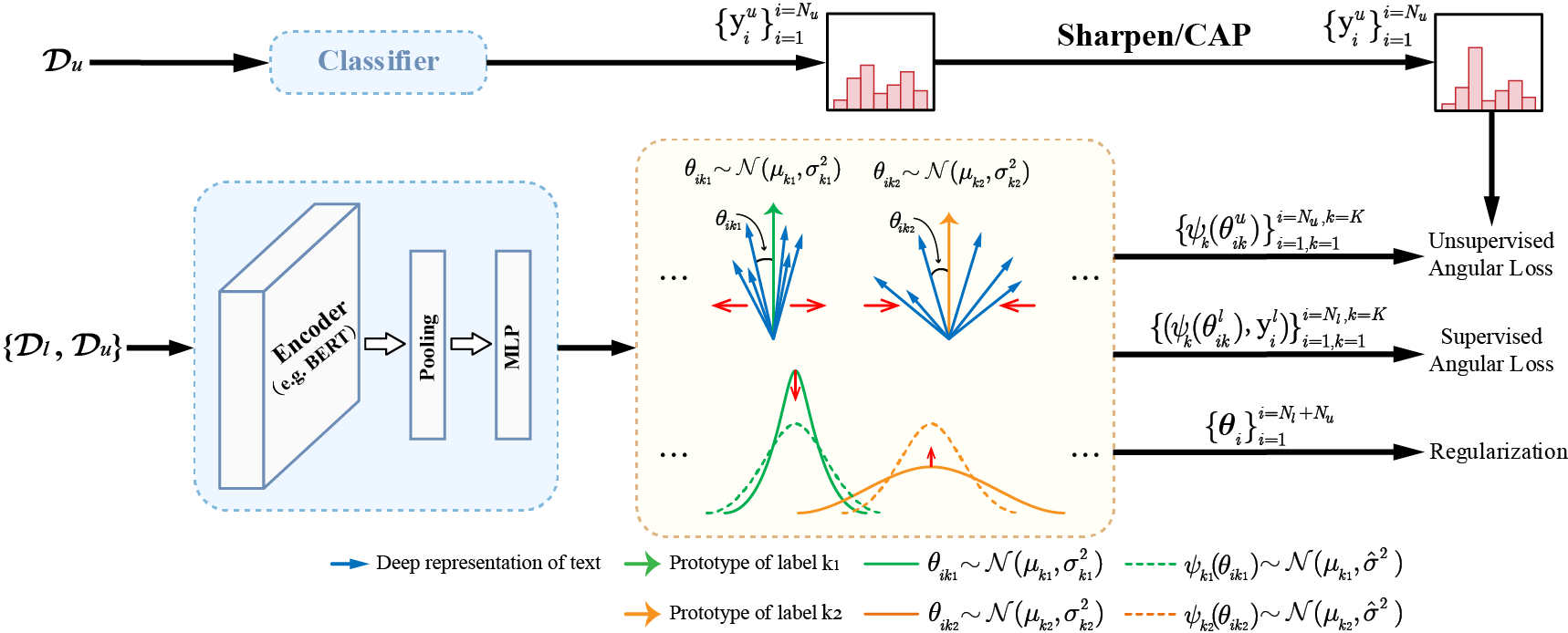}
  \centering
  \caption{Overview the framework of \baby. Specially, we employ the pseudo-labeling tricks of Sharpening and Class-distribution-Aware Pseudo-labeling (CAP) for the scenarios of multi-class classification and multi-label classification, respectively. Best viewed in color.}
  \label{SSTC}
\end{figure*}

\subsection{Overview of \baby}
\label{3.1}

Overall speaking, our \baby performs a self-training procedure for SSTC. Given a training dataset, it first trains a fine-tuned deep classifier based on the pre-trained BERT model \cite{BERT2019} with AM loss \cite{CosFace2018}. During the self-training loops, we employ the current deep classifier to predict unlabeled texts as pseudo-labels, and then update it over both labeled and pseudo-labeled texts. In particular, we develop a \textbf{B}alanced \textbf{D}eep representation \textbf{D}istribution (\textbf{BDD}) loss, aiming at more accurate pseudo-labels for unlabeled texts. The overall framework of \baby is shown in Fig.\ref{SSTC}. We now present the important details of \baby.

\vspace{5pt}
\noindent\textbf{BDD loss.}
Formally, our BDD loss is extended from the AM loss \cite{CosFace2018}. For clarity, we first describe the AM loss with respect to angles. Given a training example $(\mathbf{x}_i,\mathbf{y}_i)$, it can be formulated below:
\begin{equation}
  \label{eq3-1-1}
  \mathfrak{L}_{am}(\mathbf{x}_i,\mathbf{y}_i;\bm{\phi})= 
  -\sum_{k=1}^K y_{ik}\log \frac{e^{s(\cos(\theta_{ik})-y_{ik}m)}}{\sum_{j=1}^{K}e^{s(\cos(\theta_{ij})-y_{ij}m)}},
\end{equation}
where $\bm{\phi}$ denotes the model parameters,
\begin{equation*}
  \cos(\theta_{ik})=\frac{\mathbf{f}_i^{\top}\mathbf{W}_k}{{\norm{\mathbf{f}_i}}_2{\norm{\mathbf{W}_k}}_2},
\end{equation*}
${\norm{\cdot}}_2$ is the $\ell_2$-norm of vectors; $\mathbf{f}_i$ and $\mathbf{W}_k$ denote the deep representation of text $\mathbf{x}_i$ and the weight vector of label $k$, respectively; $\theta_{ik}$ is the angle between $\mathbf{f}_i$ and $\mathbf{W}_k$; $s$ and $m$ are the parameters used to control the rescaled norm and magnitude of cosine margin, respectively. The AM loss applies $\ell_2$ normalization on both features and weight vectors, and maximizes the decision margin in the angular space. Hence, it can force the model to maximize the inter-class variance and minimize the intra-class variance, leading to more discriminative text representations, which are beneficial for text classification. It has been used in intent detection \cite{AM1} and multilingual sentence embedding \cite{AM2}.

Reviewing Eq.\eqref{eq3-1-1}, we observe that it directly measures the loss by label angles of texts only. We consider that it corresponds to a non-optimal decision boundary in SSTC, where the difference between label angle variances is much larger than supervised learning. To alleviate this problem, we suppose that the label angles are drawn from each label-specific Gaussian distribution $\{\mathcal{N}(\mu_k,\sigma_k^2)\}_{k=1}^{k=K}$. Thanks to the properties of Gaussian distribution, we can easily transfer them into the ones with balanced variances $\{\mathcal{N}(\mu_k,\widehat{\sigma}^2)\}_{k=1}^{k=K}, \:\: \widehat \sigma^2 = \frac{\sum \nolimits_{k=1}^K\sigma_k^2}{K}$ by performing the following linear transformations to the angles:

\begin{equation}
  \label{eq3-1-2}
  \psi_k(\theta_{ik})=a_k\theta_{ik}+b_k, \; a_k=\frac{\widehat{\sigma}}{\sigma_k},\; b_k=(1-a_k)\mu_k, \; \forall k \in [K].
\end{equation}
With these linear transformations $\{\psi_k(\cdot)\}_{k=1}^{k=K}$, all angles become the samples from balanced angular distributions with the same variances, \eg $\psi_k(\theta_{ik})\sim\mathcal{N}(\mu_k,\widehat{\sigma}^2)$. Accordingly, the angular loss of Eq.\eqref{eq3-1-1} can be rewritten as the following BDD loss:
\begin{equation}
  \label{eq3-1}
  \mathfrak{L}_{bdd}(\mathbf{x}_i,\mathbf{y}_i;\bm{\phi})=
  -\sum_{k=1}^Ky_{ik}\log\frac{e^{s(\cos(\psi_k(\theta_{ik}))-y_{ik}m)}}{\sum_{j=1}^{K}e^{s(\cos(\psi_{j}(\theta_{ij}))-y_{ij}m)}}.
\end{equation}

\vspace{5pt}
\noindent\textbf{Supervised angular loss.} Applying the BDD loss $\mathfrak{L}_{bdd}$ of Eq.\eqref{eq3-1} to the labeled text set $\mathcal{D}_{l}=\{(\mathbf{x}_i^l,\mathbf{y}_i^l)\}_{i=1}^{i=N_l}$, we can formulate the following supervised angular loss:
\begin{equation}
  \label{eq3-2}
  \mathcal{L}_l(\mathcal{D}_l;\bm{\phi})=\frac{1}{N_l}\sum_{i=1}^{N_l}\mathfrak{L}_{bdd}(\mathbf{x}_i^l,\mathbf{y}_i^l;\bm{\phi}).
\end{equation}

\vspace{5pt}
\noindent\textbf{Unsupervised angular loss.} 
Under the self-training paradigm, we form the loss with unlabeled texts and pseudo-labels. Specifically, we produce the pseudo-labels according to the output probabilities of the deep classifier, which are computed by normalizing $\{\cos(\psi_k(\theta_{ik}))\}_{k=1}^{k=K}$ with the softmax function:
\begin{equation*}
  p(k|\mathbf{x}_i,\bm{\phi})=\frac{e^{\cos(\psi_k(\theta_{ik}))}}{\sum_{j=1}^{K}e^{\cos(\psi_j(\theta_{ij}))}}  \triangleq \mathbf{y}_i,\; \forall k\in[K].
\end{equation*}
For each unlabeled text $\mathbf{x}_i^u$, its classifier output probability is given by $p(k|\mathbf{x}_i^u,\widetilde{\bm{\phi}})  \triangleq \mathbf{\tilde y}_i^u $ with the \emph{fixed} copy $\widetilde{\bm{\phi}}$ of the current model parameter $\bm{\phi}$ during self-training loops. Then, we employ some pseudo-labeling tricks over $\{\mathbf{\tilde y}_i^u\}_{i=1}^{N_u}$ to generate the corresponding pseudo-labels $\{\mathbf{y}_i^u\}_{i=1}^{N_u}$ of unlabeled texts for multi-class and multi-label classification, which are described minutely in Section \ref{3.3} and \ref{3.4}, respectively. 

Applying the BDD loss of Eq.\eqref{eq3-1} to the unlabeled text set $\mathcal{D}_{u}=\{\mathbf{x}_j^u\}_{j=1}^{j=N_u}$ and pseudo-labels $\{\mathbf{y}_i^u\}_{i=1}^{N_u}$, we can formulate the following unsupervised angular loss:
\begin{equation}
  \label{eq3-3}
  \mathcal{L}_u(\mathcal{D}_u,\{\mathbf{y}_i^u\}_{i=1}^{N_u};\bm{\phi})=\frac{1}{N_u}\sum_{i=1}^{N_u}\mathfrak{L}_{bdd}(\mathbf{x}_i^u,\mathbf{y}_i^u;\bm{\phi}).
\end{equation}

\subsection{Implementations of Label Angle Variances}
\label{3.2}
In this section, we describe implementations of label angle variances. As mentioned before, what we concern is the estimations of angular distributions $\{\mathcal{N}(\mu_k,\sigma_k^2)\}_{k=1}^{k=K}$, where their draws are the angles between deep representations of texts and label prototypes denoted by $\{\mathbf{c}_k\}_{k=1}^{k=K}$. Both $\{(\mu_k,\sigma_k^2)\}_{k=1}^{k=K}$ and $\{\mathbf{c}_k\}_{k=1}^{k=K}$ are estimated over both labeled and pseudo-labeled texts during self-training loops. In the following, we describe their learning processes in more detail.

Within the framework of stochastic optimization, we update $\{(\mu_k,\sigma_k^2)\}_{k=1}^{k=K}$ and $\{\mathbf{c}_k\}_{k=1}^{k=K}$ per-epoch. For convenience, we denote $\Omega$ as the index set of labeled and unlabeled texts in one epoch, $\{\mathbf{f}_i\}_{i\in\Omega}$ and $\{\mathbf{y}_i\}_{i\in\Omega}$ as the deep representations of texts and corresponding label or pseudo-label vectors (\ie $\mathbf{y}_i^l$ or $\mathbf{y}_i^u$) in the current epoch, respectively. 

\vspace{5pt}
\noindent\textbf{Estimating label prototypes.} 
Given the current $\{\mathbf{f}_i\}_{i\in\Omega}$ and $\{\mathbf{y}_i\}_{i\in\Omega}$, we calculate the label prototypes $\{\mathbf{c}_k\}_{k=1}^{k=K}$ by the weighted average of $\{\mathbf{f}_i\}_{i\in\Omega}$, formulated below:
\begin{equation}
  \mathbf{c}_k=\frac{\sum_{i\in\Omega}y_{ik}\mathbf{f}_i}{\sum_{i\in\Omega}y_{ik}},\quad \forall k\in[K].
\end{equation}
To avoid the misleading effect of some mislabeled texts, inspired by \cite{Jialun2020}, we update $\{\mathbf{c}_k\}_{k=1}^{k=K}$ by employing the moving average with a learning rate $\gamma$:
\begin{equation*}
  \mathbf{c}_k^{(t)} \leftarrow (1-\gamma)\mathbf{c}_k^{(t)}+\gamma\mathbf{c}_k^{(t-1)}.
\end{equation*}

\vspace{5pt}
\noindent\textbf{Estimating label angle variances.} 
Given $\{\mathbf{f}_i\}_{i\in\Omega}$ and $\{\mathbf{c}_k\}_{k=1}^{k=K}$, the angles between them can be calculated by:
\begin{equation}
  \beta_{ik}=\arccos\bigl(\frac{\mathbf{f}_i^{\top}\mathbf{c}_k}{{\norm{\mathbf{f}_i}}_2{\norm{\mathbf{c}_k}}_2}\bigr),\quad \forall i\in\Omega,k\in[K].
\end{equation}
Accordingly, we can compute the estimations of $\{(\mu_k,\sigma_k^2)\}_{k=1}^{k=K}$ as follows:
\begin{equation}
  \mu_k=\frac{\sum_{i\in\Omega}y_{ik}\beta_{ik}}{\sum_{i\in\Omega}y_{ik}},\quad \sigma_k^2=\frac{\sum_{i\in\Omega}y_{ik}(\beta_{ik}-\mu_k)^2}{\sum_{i\in\Omega}y_{ik}-1}.
\end{equation}
Further, the moving average is also used for the updates below:
 \begin{align*}
  \mu_k^{(t)} &\leftarrow (1-\gamma)\mu_k^{(t)}+\gamma\mu_k^{(t-1)}, \\
  (\sigma_k^2)^{(t)} &\leftarrow (1-\gamma)(\sigma_k^2)^{(t)}+\gamma(\sigma_k^2)^{(t-1)}.
\end{align*}

\subsection{\baby for Semi-supervised Multi-class Classification}
\label{3.3}
In this section, we first describe specific designs for the multi-class classification version of \baby, including sharpening-aware pseudo-labeling and entropy regularization, and then formulate the corresponding objective function.

\vspace{5pt}
\noindent\textbf{Sharpening.}
We employ two existed sharpening-aware pseudo-labeling tricks over those classifier output probabilities $\{\mathbf{\tilde y}_i^u\}_{i=1}^{N_u}$ to produce pseudo-labels $\{\mathbf{y}_i^u\}_{i=1}^{N_u}$ of unlabeled texts in this case. 
\begin{itemize}
\item Following MixMatch \cite{MixMatch2019}, we leverage a sharpening function with a temperature $T$ over those classifier output probabilities $\{\mathbf{\tilde y}_i^u\}_{i=1}^{N_u}$:
\begin{equation}
\label{eq3-s}
  \mathbf{y}_i^u=\mathrm{Sharpen}(\mathbf{\tilde y}_i^u,T)=\frac{(\mathbf{\tilde y}_i^u)^{1/T}}{{\norm{(\mathbf{\tilde y}_i^u)^{1/T}}}_1},\:\:\:\:\:\: \forall i\in[N_u],
\end{equation}
where ${\norm{\cdot}}_1$ is the $\ell_1$-norm of vectors. When $T\rightarrow0$, the pseudo-label distribution tends to be the one-hot vector. We also linearly ramp up the weight of unlabeled instances over $[0, 1]$.
\item Following current SSL literature \cite{Dash2021,FlexMatch2021,FreeMatch2023}, we pseudo-label unlabeled texts with hard labels as $y_i^u=\arg\max(\mathbf{\tilde y}_i^u)$ and mask low-confidence instances with an adaptive confidence threshold, \eg the self-adaptive confidence threshold approach in FreeMatch \cite{FreeMatch2023}.
\end{itemize}

\vspace{5pt}
\noindent\textbf{Entropy regularization.}
We employ the conditional entropy of $p(y|\mathbf{x}_i,\bm{\phi})$ as an additional regularization term: 
\begin{align}
  \label{eq3-4}
  &\mathcal{R}_{mcc}(\mathcal{D}_l,\mathcal{D}_u;\bm{\phi})= \nonumber \\
  &\quad\; -\frac{1}{N_l+N_u}\sum_{\mathbf{x}_i\in\mathcal{D}_l,\mathcal{D}_u}\sum_{k=1}^{K}p(k|\mathbf{x}_i,\bm{\phi})\log p(k|\mathbf{x}_i,\bm{\phi}).
\end{align}
This conditional entropy regularization is introduced by \cite{EntMin2004}, and also utilized in \cite{RSTP2016,VAT2019,LSTMMIXED2019}. It also sharpens the output probability of the deep classifier.

\vspace{5pt}
\noindent\textbf{Full objective.} Combining the supervised angular loss of Eq.\eqref{eq3-2}, unsupervised angular loss of Eq.\eqref{eq3-3} and entropy regularization term of Eq.\eqref{eq3-4}, the full objective of \baby for the semi-supervised multi-class classification case can be given by:
\begin{align}
\label{eq3-5}
  &\mathcal{L}_{mcc}(\mathcal{D}_l,\mathcal{D}_u;\bm{\phi})= \nonumber \\
  &\quad\mathcal{L}_l (\mathcal{D}_l;\bm{\phi})
  +\lambda_1\mathcal{L}_u (\mathcal{D}_u,\{\mathbf{y}_i^u\}_{i=1}^{N_u};\bm{\phi}) 
  +\lambda_2\mathcal{R}_{mcc}(\mathcal{D}_l,\mathcal{D}_u;\bm{\phi}),
\end{align}
where $\lambda_1$ and $\lambda_2$ are regularization parameters. The model parameters $\bm{\phi}$ can be optimized efficiently by leveraging the stochastic gradient descent method.

\subsection{\baby for Semi-Supervised Multi-label Classification}
\label{3.4}
In this section, we first describe the multi-label classification version of \baby, including class-distribution-aware pseudo-labeling and low-rank regularization term, and then give the corresponding full objective function and the model training procedure.

\vspace{5pt}
\noindent\textbf{Class-distribution-aware pseudo-labeling.}
Due to the unknown number of ground-truth labels for each text, traditional instance-aware pseudo-labeling methods (such as sharpening) will fail in the semi-supervised multi-label classification case because it may introduce false positive labels and neglect true positive ones \cite{CAP2023}. To address this issue, we employ the class-distribution-aware pseudo-labeling (CAP) approach \cite{CAP2023}, which drives the class distribution of pseudo-labels towards the ground-truth ones, over those classifier output probabilities $\{\mathbf{\tilde y}_i^u\}_{i=1}^{N_u}$ to generate pseudo-labels $\{\mathbf{y}_i^u\}_{i=1}^{N_u}$. Specifically, we refine the pseudo-labels $\{\mathbf{y}_i^u\}_{i=1}^{N_u}$ from $\{\mathbf{\tilde y}_i^u\}_{i=1}^{N_u}$ with class-distribution-aware confidence thresholds $\{\gamma_k\}_{k=1}^K$ by:
\begin{equation}
\label{eq3-6}
    y_{ik}^u=
    \begin{cases}
        \;1\quad\text{if }\tilde{y}_{ik}^u>=\gamma_k; \\
        \\
        \;0\quad\text{otherwise},
    \end{cases}
    \:\:\:\:\:\: \forall k\in[K],\quad\forall i\in[N_u],
\end{equation}
where $\{\gamma_k\}_{k=1}^K$ are obtained by solving the equation:
\begin{equation*}
    \frac{\sum_{i=1}^{N_u}\mathbb{I}(\tilde{y}_{ik}^u\geq\gamma_k)}{N_u}=\frac{\sum_{j=1}^{N_l}y_{jk}^l}{N_l}, \:\:\:\:\:\: \forall k\in[K],
\end{equation*}
and $\mathbb{I}(z)=1$ if $z$ is true or $0$ otherwise. 

\vspace{5pt}
\noindent\textbf{Regularization with low-rank constraint.}
Considering the label correlation in multi-label classification, we introduce the following low-rank constraint over the weight vectors $\{\mathbf{W}_k\}_{k=1}^K$ of the classifier for all labels as an additional regularization term:
\begin{equation}
\label{eq3-7}
    \mathcal{R}_{mlc}(\mathcal{D}_l,\mathcal{D}_u;\bm{\phi})=\lVert \mathbf{W} \rVert_{*},
\end{equation}
where $\lVert\cdot\rVert_{*}$ is the nuclear norm.

\vspace{5pt}
\noindent\textbf{Full objective and model training.}
Combining the supervised angular loss of Eq.\eqref{eq3-2}, unsupervised angular loss of Eq.\eqref{eq3-3}, and low-rank regularization term of Eq.\eqref{eq3-7}, the full objective of \baby for the semi-supervised multi-label classification case can be formulated below:
\begin{align}
\label{eq3-8}
  &\mathcal{L}_{mlc}(\mathcal{D}_l,\mathcal{D}_u;\bm{\phi})= \nonumber \\
  &\quad \mathcal{L}_l (\mathcal{D}_l;\bm{\phi})
  +\lambda_1\mathcal{L}_u (\mathcal{D}_u,\{\mathbf{y}_i^u\}_{i=1}^{N_u};\bm{\phi}) 
  +\lambda_3\mathcal{R}_{mlc}(\mathcal{D}_l,\mathcal{D}_u;\bm{\phi}),
\end{align}
where $\lambda_1$ and $\lambda_3$ are regularization parameters. We optimize the model parameters $\bm{\phi}$ by minimizing the objective of Eq.\eqref{eq3-8}. For efficient optimization, we employ the ADMM method \cite{ADMM2011} and reformulate Eq.\eqref{eq3-8} by incorporating an auxiliary parameter $\widehat{\mathbf{W}}$ as follows:
\begin{align}
\label{eq3-9}
  &\mathop{\textbf{min}}\limits_{\bm{\phi},\widehat{\mathbf{W}}}\mathcal{L}_l (\mathcal{D}_l;\bm{\phi})
  +\lambda_1\mathcal{L}_u (\mathcal{D}_u,\{\mathbf{y}_i^u\}_{i=1}^{N_u};\bm{\phi}) 
  +\lambda_3\lVert \widehat{\mathbf{W}} \rVert_{*}, \nonumber \\
  &\quad\textbf{s.t.}\;\mathbf{W}=\widehat{\mathbf{W}},
\end{align}
whose augmented Lagrangian form is given by:
\begin{align}
\label{eq3-10}
  &\mathop{\textbf{min}}\limits_{\bm{\phi},\widehat{\mathbf{W}},\bm{\Theta}}\mathcal{L}_l (\mathcal{D}_l;\bm{\phi})  
  +\lambda_1\mathcal{L}_u (\mathcal{D}_u,\{\mathbf{y}_i^u\}_{i=1}^{N_u};\bm{\phi}) \nonumber \\
  &\quad\quad\quad+\lambda_3\lVert \widehat{\mathbf{W}} \rVert_{*}+\frac{\tau}{2}\Bigl\lVert \widehat{\mathbf{W}}-\mathbf{W}+\frac{\bm{\Theta}}{\tau} \Bigr\rVert_F^2,
\end{align}
where $\lVert\cdot\rVert_F^2$ is the squared Frobenius norm, $\bm{\Theta}$ a Lagrange parameter, $\tau$ a penlty parameter. The parameters $\{\bm{\phi},\widehat{\mathbf{W}},\bm{\Theta}\}$ can be efficiently updated by the following alternative optimization procedure:
\begin{enumerate}
\item Fixing $\{\widehat{\mathbf{W}},\bm{\Theta}\}$ as constants, the optimization with respect to $\bm{\phi}$ can be compactedly stated as follows:
\begin{equation*}
  \mathop{\textbf{min}}\limits_{\bm{\phi}}\mathcal{L}_l (\mathcal{D}_l;\bm{\phi})
  +\lambda_1\mathcal{L}_u (\mathcal{D}_u,\{\mathbf{y}_i^u\}_{i=1}^{N_u};\bm{\phi}) 
  +\frac{\tau}{2}\Bigl\lVert \widehat{\mathbf{W}}-\mathbf{W}+\frac{\bm{\Theta}}{\tau} \Bigr\rVert_F^2,
\end{equation*}
and $\bm{\phi}$ can be updated with the stochastic gradient descent method.
\item When fixing $\{\bm{\phi},\bm{\Theta}\}$ as constants, the optimization with respect to $\widehat{\mathbf{W}}$ can be given by:
\begin{equation*}
  \mathop{\textbf{min}}\limits_{\widehat{\mathbf{W}}}\frac{\lambda_3}{\tau}\lVert \widehat{\mathbf{W}} \rVert_{*}+\frac{1}{2}\Bigl\lVert \widehat{\mathbf{W}}-\mathbf{W}+\frac{\bm{\Theta}}{\tau} \Bigr\rVert_F^2.
\end{equation*}
Following \cite{SVTA2010}, $\widehat{\mathbf{W}}$ can be directly solved by:
\begin{equation*}
    \widehat{\mathbf{W}}\leftarrow D_{\frac{\lambda_3}{\tau}}(\mathbf{W}-\frac{\bm{\Theta}}{\tau}),
\end{equation*}
where $D_{\frac{\lambda_3}{\tau}}(\cdot)$ is the singular value thresholding.
\item By holding $\{\bm{\phi},\widehat{\mathbf{W}}\}$ fixed, following \cite{ADMM2011}, $\bm{\Theta}$ can be updated by
$\bm{\Theta}\leftarrow\bm{\Theta}+\tau(\widehat{\mathbf{W}}-\mathbf{W})$.
\end{enumerate}

%% file: Sec_Experiment.tex
\section{Experiment}
\label{4}
In this section, we examine the effectiveness of our \baby on both semi-supervised multi-class and semi-supervised multi-label text classification tasks.

\subsection{Experimental Settings}

\vspace{5pt}
\noindent\textbf{Datasets.}
We employ 3 benchmark datasets for multi-class text classification: \emph{AG News} \cite{Data2015}, \emph{Yelp} \cite{Data2015}, and \emph{Yahoo} \cite{Yahoo2008}; and 3 benchmark datasets for multi-label text classification: Ohsumed\footnote{\url{http://disi.unitn.it/moschitti/corpora.htm}}, AAPD\footnote{\url{https://github.com/lancopku/SGM}} \cite{SGM2018}, RCV1-V2\footnote{\url{http://www.ai.mit.edu/projects/jmlr/papers/volume5/lewis04a/lyrl2004_rcv1v2_README.htm}} \cite{RCV12004}. For Ohsumed, we leverage the 70\%/30\% random train/test split. For AAPD and RCV1-V2, we utilize the corresponding train-test split proposed in \cite{SGM2018} and \cite{RCV12004}, respectively.
For all datasets, we form the labeled training set, unlabeled training set, and development set by randomly drawing from the corresponding original training datasets, and utilize the original test sets for prediction evaluation. The dataset statistics and split information are described in Table \ref{table5-1}. 

\vspace{5pt}
\noindent\textbf{Baselines.}
We choose 8 existing baselines for multi-class text classification, described below.
\begin{itemize}
\item\textbf{NB+EM} \cite{NBMEM2000}: A semi-supervised multi-class text classification method combining a Naive Bayes classifier (NB) and  Expectation-Maximization (EM). In experiments, we pre-process texts following \cite{VAM2019} and use tf-idfs as the representations of texts.
\item\textbf{BERT+CE} \cite{BERT2019}: A supervised multi-class text classification method built on the pre-trained BERT-based-uncased model\footnote{\url{https://pypi.org/project/pytorch-transformers/}} and fine-tuned with the supervised softmax with cross-entropy loss on labeled texts.
\item\textbf{BERT+AM}: A semi-supervised multi-class text classification method built on the pre-trained BERT-based-uncased\footnotemark[4] and fine-tuned following the self-training spirit with the AM loss on both labeled and unlabeled texts. 
\item\textbf{VAMPIRE} \cite{VAM2019}: A semi-supervised multi-class text classification method based on variational pre-training. The code is available on the net.\footnote{\url{https://github.com/allenai/vampire}} In experiments, the default parameters are utilized.
\item\textbf{VAT} \cite{VAT2019}: A semi-supervised multi-class text classification method based on virtual adversarial training. [parameter configuration: perturbation size $\epsilon=5.0$, regularization coefficient $\alpha=1.0$, hyperparameter for finite difference $\xi=0.1$]
\item\textbf{UDA} \cite{UDA2020}: A semi-supervised multi-class text classification method based on unsupervised data augmentation with back translation. The code is available on the net.\footnote{\url{https://github.com/google-research/uda}} In experiments, we utilize the default parameters, and generate the augmented unlabeled data by using FairSeq\footnote{\url{https://github.com/pytorch/fairseq}} with German as the intermediate language.
\item\textbf{FreeMatch} \cite{FreeMatch2023}: A semi-supervised multi-class text classification method based on a self-adaptive confidence threshold strategy. The code is available on the net.\footnote{\url{https://github.com/microsoft/Semi-supervised-learning}} In experiments, the default parameters are used, and the strong augmentation is the same with UDA.
\item\textbf{MetaExpert} \cite{MetaExpert2025}: A semi-supervised multi-class text classification method based on a dynamic expert assignment strategy. The code is available on the net.\footnote{\url{https://github.com/YaxinHou/Meta-Expert}} In experiments, the default parameters are used.
\end{itemize}

\begin{table}[t]
 \footnotesize
  \caption{Statistics of datasets. $K$: the number of class labels. $N$: the number of original training texts. \emph{\#Dev}: the number of development texts. \emph{\#Test}: the number of texts for testing. In addition, we report the average number of category labels per document (\emph{Avg.Labels}) for multi-label benchmarks.}
  \renewcommand\arraystretch{1.1}
  \label{table5-1}
  \begin{center}
      \setlength{\tabcolsep}{3pt}{
        \begin{tabular}{c|c|cccccc}
          \Xhline{1.5pt}
          Type &Dataset     &$K$ &$N$    &\#Dev  &\#Test &Avg.Labels\\
          \Xhline{1pt}
         \multirow{3}{*}{\tabincell{c}{Multi-\emph{class}\\classification}} 
         &\emph{AG News}  &4  &120,000     &8,000  &7,600    &-- \\
          ~ &\emph{Yelp}    &5  &650,000      &10,000   &50,000  &--  \\
          ~ &\emph{Yahoo}    &10	&700,000   &20,000	  &60,000  &-- \\
          \Xhline{1pt}
          \multirow{3}{*}{\tabincell{c}{Multi-\emph{label}\\classification}} 
          &\emph{Ohsumed}    &23 &24,089  &2,000 &10,300  &1.66 \\
          ~ &\emph{AAPD}       &54  &54,840     &2,000  &1,000   &2.41  \\
          ~ &\emph{RCV1-V2}    &103  &23,149      &2,000   &781,265 &3.24   \\
          \Xhline{1.5pt}
        \end{tabular}}
  \end{center}
\end{table}


We also choose 5 existing baselines for multi-label text classification, described below.
\begin{itemize}
    \item\textbf{BERT+BCE} \cite{BERT2019}: A supervised multi-label text classification method built on the pre-trained BERT-based-uncased model\footnotemark[4] and fine-tuned with the supervised binary cross entropy (BCE) loss on labeled texts.
    \item\textbf{BERT+AM}: Same with the above-mentioned semi-supervised multi-class text classification method BERT+AM, but applied to the multi-label text classification scenario with the class-aware pseudo-labeling approach.
    \item\textbf{SDRL} \cite{DRML2020,DRML2021}: A semi-supervised multi-label classification method based on feature-level and label-level correlations. The code is available on the net.\footnote{\url{https://github.com/wanglichenxj/Dual-Relation-Semi-supervised-Multi-label-Learning}}
    \item\textbf{CAP} \cite{CAP2023}: A semi-supervised multi-label classification method based on the class-aware pseudo-labeling and the deep self-training strategy with asymmetric loss. The code is available on the net.\footnote{\url{https://github.com/milkxie/SSMLL-CAP}}
    \item\textbf{MetaExpert} \cite{MetaExpert2025}: Same with the above-mentioned semi-supervised multi-class text classification method MetaExpert, but applied to the multi-label text classification scenario with asymmetric loss. In experiments, the default parameters are used.
\end{itemize}

\begin{table*}[htp]
  \centering
  \caption{Experimental results of Micro-F1 and Macro-F1 by varying the number of labeled texts $N_l$ on multi-class benchmarks. The best results are highlighted in boldface.}
  \renewcommand\arraystretch{1.05}
  \label{table5-2}
\resizebox{0.98\linewidth}{!}{
  \begin{tabular}{c|c|c||p{35pt}<{\centering}p{35pt}<{\centering}p{55pt}<{\centering}p{55pt}<{\centering}p{35pt}<{\centering}p{35pt}<{\centering}p{35pt}<{\centering}p{35pt}<{\centering}p{45pt}<{\centering}p{45pt}<{\centering}p{45pt}<{\centering}}
    \Xhline{1.5pt}
    Metric &Dataset  &$N_l$  &\textbf{NB+EM}  &\textbf{BERT+CE} &\textbf{BERT+AM\textsc{-s}} &\textbf{BERT+AM\textsc{-f}} &\textbf{VAMPIRE} &\textbf{VAT} &\textbf{UDA} &\textbf{FreeMatch} &\textbf{MetaExpert} &\textbf{\babys} &\textbf{\babyf} \\
    \Xhline{0.5pt}
    \hline
    \hline
    \Xhline{0.5pt}
    \multirow{9}{*}{Micro-F1} &\multirow{3}{*}{\emph{AG News}}    &100  &{0.834}    &{0.839} &{0.856} &{0.870} &{0.705} &{0.868} &{0.855} &{0.862} &{0.867}  &{0.872} &\textbf{0.876} \\
    ~&~&1,000   &{0.855}  &{0.878} &{0.879} &{0.886} &{0.833}  &{0.886}  &{0.883} &{0.882} &{0.884} &\textbf{0.889} &\textbf{0.889}  \\
    ~&~&10,000  &{0.874}  &{0.905} &{0.901} &{0.903} &{0.876}  &{0.898}  &{0.906}  &{0.915} &{0.914} &{0.907} &\textbf{0.917}  \\
    \cline{2-14}
    ~&\multirow{3}{*}{\emph{Yelp}}       &100  &{0.300}   &{0.344} &{0.399} &{0.485} &{0.227} &{0.244} &{0.387} &{0.491} &{0.454} &{0.417} &\textbf{0.496} \\
    ~&~&1,000   &{0.355} &{0.538} &{0.544} &{0.567} &{0.476}  &{0.551}  &{0.554} &{0.565}  &{0.585} &{0.552}  &\textbf{0.589} \\
    ~&~&10,000  &{0.404} &{0.583}  &{0.574} &{0.596} &{0.551}  &{0.566}  &{0.580} &{0.618} &{0.616}  &{0.583}  &\textbf{0.619} \\
    \cline{2-14}
    ~&\multirow{3}{*}{\emph{Yahoo}}  &100 &{0.529} &{0.564} &{0.589} &{0.641} &{0.389} &{0.534} &{0.576} &{0.672} &{0.667} &{0.618} &\textbf{0.688} \\
    ~&~&1,000  &{0.624}  &{0.676} &{0.679} &{0.694} &{0.547}  &{0.685}  &{0.672} &{0.694} &{0.703}  &{0.687} &\textbf{0.706}  \\
    ~&~&10,000  &{0.659} &{0.713} &{0.706} &{0.722} &{0.644}  &{0.701}  &{0.707} &{0.717} &{0.722}  &{0.713}  &\textbf{0.725} \\
    \Xhline{1.2pt}
    \multirow{9}{*}{Macro-F1} &\multirow{3}{*}{\emph{AG News}}    &100  &{0.833} &{0.840} &{0.856} &{0.870} &{0.698} &{0.867} &{0.855} &{0.862} &{0.867} &{0.872} &\textbf{0.876} \\
    ~&~&1,000   &{0.855}  &{0.878} &{0.879} &{0.886} &{0.833}  &{0.886}  &{0.883} &{0.882} &{0.883} &{0.889}  &\textbf{0.889}  \\
    ~&~&10,000  &{0.873}  &{0.905} &{0.900} &{0.903} &{0.876}  &{0.897}  &{0.906} &{0.915} &{0.914}  &{0.907} &\textbf{0.917}  \\
    \cline{2-14}
    ~&\multirow{3}{*}{\emph{Yelp}}       &100  &{0.250} &{0.324} &{0.371} &{0.468} &{0.144} &{0.197} &{0.357} &{0.490} &{0.457} &{0.403} &\textbf{0.492} \\
    ~&~&1,000   &{0.329} &{0.532} &{0.535} &{0.558} &{0.476}  &{0.548}  &{0.550} &{0.554} &{0.581}  &{0.550} &\textbf{0.591}  \\
    ~&~&10,000  &{0.397} &{0.586} &{0.562} &{0.597} &{0.553}  &{0.569}  &{0.576} &{0.617} &{0.616}  &{0.586} &\textbf{0.618}  \\
    \cline{2-14}
    ~&\multirow{3}{*}{\emph{Yahoo}}  &100 &{0.489} &{0.550} &{0.573} &{0.633} &{0.356} &{0.542} &{0.567} &{0.672} &{0.655} &{0.595} &\textbf{0.684} \\
    ~&~&1,000  &{0.616}  &{0.671} &{0.672} &{0.689} &{0.545}  &{0.675}  &{0.666} &{0.692} &{0.695}  &{0.680} &\textbf{0.702}  \\
    ~&~&10,000  &{0.653} &{0.708} &{0.695} &{0.714} &{0.644}  &{0.697}  &{0.704} &{0.714} &{0.718}  &{0.709}  &\textbf{0.722} \\
    \Xhline{1pt}
    \multicolumn{3}{c||}{Average Rank} &{10.2} &{7.5} &{7.4} &{3.7} &{10.7} &{7.2} &{6.7} &{3.5} &{3.2} &{4.2} &\textbf{1.0} \\
    \Xhline{1.5pt}
  \end{tabular}
  }
\end{table*}

\vspace{5pt}
\noindent\textbf{Implementation details.} 
Note that, as mentioned in Section \ref{3.3}, for semi-supervised multi-class classification, we implement two versions of \baby with different sharpening approaches: the sharpening function Eq.\eqref{eq3-s} with the linear ramp-up of weights of unlabeled texts denoted by \textbf{\babys}, and the self-adaptive confidence threshold with hard labels and strong augmentation as FreeMatch \cite{FreeMatch2023} denoted by \textbf{\babyf}. And the corresponding variants of BERT+AM are denoted by BERT+AM\textsc{-s} and BERT+AM\textsc{-f} respectively.
For \baby (\textsc{-s} and \textsc{-f}), BERT+CE, BERT+AM (\textsc{-s} and \textsc{-f}), VAT, UDA, FreeMatch, MetaExpert, BERT+BCE, SDRL and CAP, we utilize BERT-based-uncased tokenizer to tokenize texts; average pooling over BERT-based-uncased model as text encoder to encode texts; and a two-layer MLP, whose hidden size and activation function are 128 and tanh respectively, as the classifier to predict labels. We set the max sentence length as 256 and retain the first 256 tokens for texts exceeding the length limit. For optimization, we utilize AdamW \cite{AdamW2019} optimizer with learning rates of 1e-5 for BERT encoder and 1e-3 for MLP classifier. For BERT+CE and BERT+BCE, we set the batch size of labeled texts as 8. For \baby (\textsc{-s} and \textsc{-f}), BERT+AM (\textsc{-s} and \textsc{-f}), VAT, UDA, FreeMatch, MetaExpert, BERT+BCE, SDRL and CAP, the batch sizes of labeled and unlabeled texts are 4 and 8, respectively. For all datasets, we iterate 50 epochs, where each one contains 1000 inner loops. Specifically, on all multi-label benchmarks, for \baby, BERT+AM, CAP and MetaExpert, we warm up the encoder and classifier on labeled texts (AM loss for \baby and BERT+AM, Asymmetric loss for CAP and MetaExpert) with numbers of epochs and batch size as 5 and 8. All experiments are performed on a Linux server with NVIDIA GeForce RTX 4090 GPUs. 

\begin{table*}[!ht]
  \centering
  \caption{Experimental results of Micro-F1 and Macro-F1 by varying the number of unlabeled texts $N_u$ on multi-class benchmarks.}
  \renewcommand\arraystretch{1.05}
  \label{table5-3}

\resizebox{0.98\linewidth}{!}{
  \begin{tabular}{c|c|c||p{30pt}<{\centering}p{50pt}<{\centering}p{50pt}<{\centering}p{30pt}<{\centering}p{30pt}<{\centering}p{30pt}<{\centering}p{30pt}<{\centering}p{50pt}<{\centering}p{50pt}<{\centering}p{50pt}<{\centering}}
    \Xhline{1.5pt}
    Metric &Dataset  &$N_u$  &\textbf{NB+EM} &\textbf{BERT+AM\textsc{-s}} &\textbf{BERT+AM\textsc{-f}} &\textbf{VAMPIRE} &\textbf{VAT} &\textbf{UDA} &\textbf{FreeMatch} &\textbf{MetaExpert} &\textbf{\babys}  &\textbf{\babyf} \\
    \Xhline{0.5pt}
    \hline
    \hline
    \Xhline{0.5pt}
    \multirow{12}{*}{Micro-F1} &\multirow{4}{*}{\emph{AG News}} &0  &{0.668} &{0.844} &{0.798}   &{--} &{0.846} &{0.839} &\textbf{0.859} &{0.842} &{0.844}  &{0.750} \\
    ~&~&200   &{0.696} &{0.855} &{0.847}  &{0.329}  &{0.850}  &{0.844}  &{0.854} &{0.848} &\textbf{0.857}  &{0.835}  \\
    ~&~&2,000  &{0.752} &{0.856} &{0.866}  &{0.421}  &{0.870}  &{0.853}  &{0.845} &{0.859} &{0.863}  &\textbf{0.874}   \\
    ~&~&20,000  &{0.834}  &{0.856} &{0.870} &{0.705}  &{0.868}  &{0.855}   &{0.862}  &{0.867} &{0.872}  &\textbf{0.876} \\
    \cline{2-13}
    ~&\multirow{4}{*}{\emph{Yelp}}       &0  &{0.317}  &{0.381} &{0.377}  &{--} &{0.341} &{0.344} &\textbf{0.463} &{0.437} &{0.395} &{0.373} \\
    ~&~&200   &{0.307} &{0.385} &\textbf{0.420}  &{0.238}  &{0.299}  &{0.397}  &{0.299} &{0.410} &{0.403}  &\textbf{0.420}  \\
    ~&~&2,000  &{0.302} &{0.393} &\textbf{0.460}  &{0.211}  &{0.294}  &{0.379}  &{0.434} &{0.451} &{0.417}   &{0.457} \\
    ~&~&20,000  &{0.300} &{0.399} &{0.485}  &{0.227}  &{0.244}  &{0.387}   &{0.491} &{0.454} &{0.417} &\textbf{0.496} \\
    \cline{2-13}
    ~&\multirow{4}{*}{\emph{Yahoo}}       &0  &{0.312} &{0.581} &\textbf{0.606}   &{--} &{0.557} &{0.564} &{0.604} &{0.610} &{0.590} &{0.598} \\
    ~&~&400   &{0.318} &{0.582} &{0.599}  &{0.162}  &{0.519}  &{0.508}  &{0.592} &{0.590} &{0.593}  &\textbf{0.605}  \\
    ~&~&4,000  &{0.442} &{0.584} &{0.630}  &{0.221}  &{0.523}  &{0.559}  &{0.593} &{0.630} &{0.598}  &\textbf{0.645}  \\
    ~&~&40,000  &{0.529} &{0.589} &{0.641}  &{0.389}  &{0.534}  &{0.576}   &{0.672} &{0.667} &{0.618} &\textbf{0.688} \\
    \Xhline{1.2pt}
    \multirow{12}{*}{Macro-F1} &\multirow{4}{*}{\emph{AG News}}    &0  &{0.667}  &{0.843} &{0.795}   &{--} &{0.845} &{0.840} &\textbf{0.858} &{0.840} &{0.843} &{0.751} \\
    ~&~&200   &{0.695} &{0.855} &{0.847}  &{0.219}  &{0.850}  &{0.843}  &{0.854} &{0.847} &\textbf{0.857}  &{0.834}  \\
    ~&~&2,000  &{0.751} &{0.855} &{0.867}  &{0.341}  &{0.870}  &{0.852}  &{0.845} &{0.859} &{0.864}  &\textbf{0.873}   \\
    ~&~&20,000  &{0.833} &{0.856} &{0.870}  &{0.698}  &{0.867}  &{0.855}   &{0.862} &{0.867}  &{0.872} &\textbf{0.876}  \\
    \cline{2-13}
    ~&\multirow{4}{*}{\emph{Yelp}}       &0  &{0.316} &{0.368} &{0.379}   &{--} &{0.256} &{0.324} &\textbf{0.467} &{0.422} &{0.385} &{0.376} \\
    ~&~&200   &{0.279} &{0.370} &{0.406}  &{0.161}  &{0.278}  &{0.344}  &{0.252} &{0.407} &{0.372}  &\textbf{0.413}  \\
    ~&~&2,000  &{0.286} &{0.379} &{0.450}  &{0.124}  &{0.287}  &{0.362}  &{0.405} &{0.432} &{0.380}  &\textbf{0.455}  \\
    ~&~&20,000  &{0.250} &{0.371} &{0.468}  &{0.144}  &{0.197}  &{0.357}  &{0.490} &{0.457}  &{0.403} &\textbf{0.492} \\
    \cline{2-13}
    ~&\multirow{4}{*}{\emph{Yahoo}}       &0  &{0.303} &{0.567} &{0.604}   &{--} &{0.562} &{0.550} &\textbf{0.607} &{0.599} &{0.585} &{0.595} \\
    ~&~&400   &{0.301} &{0.571} &{0.597}  &{0.074}  &{0.521}  &{0.500}  &{0.589} &{0.580} &{0.586}  &\textbf{0.601}  \\
    ~&~&4,000  &{0.420} &{0.574} &{0.624}  &{0.175}  &{0.524}  &{0.550}  &{0.587} &{0.613} &{0.590}  &\textbf{0.636}  \\
    ~&~&40,000  &{0.489} &{0.573} &{0.633}  &{0.356}  &{0.542}  &{0.567}  &{0.672} &{0.655}  &{0.595} &\textbf{0.684} \\
    \Xhline{1pt}
    \multicolumn{3}{c||}{Average Rank} &{8.7} &{5.4} &{3.3}  &{10.0} &{6.4} &{7.0} &{3.8} &{3.6} &{3.7} &\textbf{2.8} \\
    \Xhline{1.5pt}
  \end{tabular}
  }
\end{table*}

\vspace{5pt}
\noindent\textbf{Parameter settings.} For \baby, in our experiments, its parameters are mostly set as: $\lambda_1=1.0$, $\lambda_2=1.0$, $s=1.0$, $m=0.01$ for \babys, and $\lambda_1=1.0$, $\lambda_2=0.001$, $s=20.0$, $m=0.3$ for \babyf on multi-class benchmarks; $\lambda_1=1.0$, $\lambda_3=0.001$, $s=20.0$, $m=0.3$ on multi-label benchmarks. Specifically, on \emph{Yelp} we set $m=0.3$ for \babys and $m=0.15$ for \babyf. For the sharpening temperature $T$ of \babys, we set $0.5$ for \emph{AG News} and \emph{Yahoo}, $0.3$ for \emph{Yelp}. The learning rate $\gamma$ of label prototypes and label angle variances is set to $0.1$ for multi-class benchmarks and $0.001$ for multi-label benchmarks. Furthermore, we perform the exponential moving average (EMA) technique for the model parameter $\bm{\phi}$ with a decay of 0.999.

\vspace{5pt}
\noindent\textbf{Metrics.} For semi-supervised multi-class text classification, we utilize two metrics of Micro-F1 and Macro-F1, which are two different types of averaged F1 scores. For semi-supervised multi-label text classification, we employ two label-based classification metrics of Micro-F1 and Macro-F1, and two example-based ranking metrics of Ranking Loss and average precision (AP) \cite{MLL2014}.
In experiments, we employ the implementation of all metrics
in the public Scikit-Learn \cite{SKLEARN2011} tool.\footnote{\url{https://scikit-learn.org/stable/}}
For all methods, we perform each method with five random seeds, and report the average scores.

\subsection{Results of Semi-Sueprvised Multi-Class Classification}
\label{4.2}

\subsubsection{Varying Number of Labeled Texts}
We first evaluate the classification performance of \baby with different amounts of labeled texts. For all methods, we conduct the experiments by varying the number of labeled texts $N_l$ over the set $\{100,1000,10000\}$ with the number of unlabeled texts $N_u=20000$ for \emph{AG News} and \emph{Yelp}, and $N_u=40000$ for \emph{Yahoo}. The classification results of both Micro-F1 and Macro-F1 over all datasets are shown in Table \ref{table5-2}, in which the best scores among all comparing baselines are highlighted in boldface.

\begin{table*}[htp]
  \centering
  \caption{Experimental results of Micro-F1 and Macro-F1 by varying the number of labeled texts $N_l$ on multi-label benchmarks. The best results are highlighted in boldface.}
  \renewcommand\arraystretch{1.05}
  \label{table5-2-1}

\resizebox{0.7\linewidth}{!}{
  \begin{tabular}{c|c|c||p{40pt}<{\centering}p{40pt}<{\centering}p{40pt}<{\centering}p{40pt}<{\centering}p{40pt}<{\centering}p{40pt}<{\centering}}
    \Xhline{1.5pt}
    Metric &Dataset  &$N_l$ &\textbf{BERT+BCE} &\textbf{BERT+AM} &\textbf{SDRL} &\textbf{CAP} &\textbf{MetaExpert} &\textbf{\baby} \\
    \Xhline{0.5pt}
    \hline
    \hline
    \Xhline{0.5pt}
    \multirow{9}{*}{Micro-F1} 
    &\multirow{3}{*}{\emph{Ohsumed}}  &100  &{0.278} &{0.421}  &{0.311} &{0.380} &{0.418} &\textbf{{0.447}} \\
    ~&~&1,000   &{0.583} &{0.604}  &{0.397} &{0.571} &{0.558} &\textbf{{0.636}}   \\
    ~&~&10,000   &{0.695} &{0.721}  &{0.443} &{0.645} &{0.644} &\textbf{{0.726}}   \\
    \cline{2-9}
    ~&\multirow{3}{*}{\emph{AAPD}}    &200  &{0.456} &{0.543}  &{0.361} &{0.497} &{0.525} &\textbf{{0.554}} \\
    ~&~&2,000    &{0.623} &{0.637}  &{0.416} &{0.625} &{0.601} &\textbf{{0.651}}   \\
    ~&~&20,000  &{0.678} &\textbf{{0.707}}  &{0.440} &{0.627} &{0.610} &\textbf{{0.707}}   \\
    \cline{2-9}
    ~&\multirow{3}{*}{\emph{RCV1-V2}}       &100   &{0.516} &{0.649} &{0.479} &{0.568} &{0.559} &\textbf{{0.655}} \\
    ~&~&1,000    &{0.784} &{0.815} &{0.585} &{0.752} &{0.687} &\textbf{{0.821}}   \\
    ~&~&10,000   &{0.840} &{0.864}  &{0.650} &{0.774} &{0.736} &\textbf{{0.876}}   \\
    
    \Xhline{1.2pt}
    \multirow{9}{*}{Macro-F1} 
    &\multirow{3}{*}{\emph{Ohsumed}}  &100  &{0.131} &{0.298}  &{0.107} &{0.252} &{0.300} &\textbf{{0.308}} \\
    ~&~&1,000   &{0.512} &{0.552} &{0.190} &{0.512} &{0.512} &\textbf{{0.562}}   \\
    ~&~&10,000   &{0.657} &{0.666} &{0.219} &{0.615} &{0.611} &\textbf{{0.678}}   \\
    \cline{2-9}
    ~&\multirow{3}{*}{\emph{AAPD}}    &200   &{0.212} &{0.342} &{0.070} &{0.315} &{0.328} &\textbf{{0.356}} \\
    ~&~&2,000    &{0.420} &{0.466} &{0.102} &{0.469} &{0.440} &\textbf{{0.483}}   \\
    ~&~&20,000   &{0.465} &{0.538} &{0.125} &{0.496} &{0.457} &\textbf{{0.553}}   \\
    \cline{2-9}
    ~&\multirow{3}{*}{\emph{RCV1-V2}}       &100   &{0.127} &{0.270} &{0.082} &{0.224} &{0.235} &\textbf{{0.277}} \\
    ~&~&1,000    &{0.455} &{0.524} &{0.159} &{0.483} &{0.403} &\textbf{{0.532}}   \\
    ~&~&10,000   &{0.611} &{0.645} &{0.211} &{0.550} &{0.451} &\textbf{{0.646}}   \\
    
    \Xhline{1pt}
    \multicolumn{3}{c||}{Average Rank}  &{4.0} &{2.1} &{5.9} &{3.6} &{4.2} &\textbf{{1.0}} \\
    \Xhline{1.5pt}
  \end{tabular}
  }
\end{table*}

\begin{table*}[!ht]
  \centering
  \caption{Experimental results of the ranking loss and average precision (AP) by varying the number of labeled texts $N_l$ on multi-label benchmarks. The best results are highlighted in boldface.}
  \renewcommand\arraystretch{1.05}
  \label{table5-2-2}

\resizebox{0.7\linewidth}{!}{
  \begin{tabular}{c|c|c||p{40pt}<{\centering}p{40pt}<{\centering}p{40pt}<{\centering}p{40pt}<{\centering}p{40pt}<{\centering}p{40pt}<{\centering}}
    \Xhline{1.5pt}
    Metric &Dataset  &$N_l$ &\textbf{BERT+BCE} &\textbf{BERT+AM}  &\textbf{SDRL} &\textbf{CAP} &\textbf{MetaExpert} &\textbf{\baby} \\
    \Xhline{0.5pt}
    \hline
    \hline
    \Xhline{0.5pt}
    \multirow{9}{*}{Ranking Loss} 
    &\multirow{3}{*}{\emph{Ohsumed}}  &100  &{0.246} &{0.208} &{0.249} &{0.199} &{0.192} &\textbf{{0.184}} \\
    ~&~&1,000   &{0.094} &{0.099}  &{0.205} &{0.089} &{0.091} &\textbf{{0.083}}   \\
    ~&~&10,000   &\textbf{{0.045}} &{0.049}  &{0.172} &\textbf{0.045} &{0.049} &{0.049}   \\
    \cline{2-9}
    ~&\multirow{3}{*}{\emph{AAPD}}    &200  &{0.121} &{0.110}  &{0.196} &\textbf{{0.096}} &\textbf{0.096} &{0.105} \\
    ~&~&2,000    &{0.063} &{0.061}  &{0.160} &\textbf{{0.051}} &{0.073} &{0.056}   \\
    ~&~&20,000  &{0.037} &\textbf{{0.035}}  &{0.137} &{0.037} &{0.041} &{0.038}  \\
    \cline{2-9}
    ~&\multirow{3}{*}{\emph{RCV1-V2}}       &100   &{0.131} &{0.047}  &{0.090} &{0.056} &\textbf{0.046} &{0.047} \\
    ~&~&1,000    &{0.018} &{0.015}  &{0.055} &{0.015} &{0.015} &\textbf{{0.014}}   \\
    ~&~&10,000   &\textbf{{0.008}} &\textbf{{0.008}}  &{0.041} &\textbf{0.008} &{0.012} &\textbf{{0.008}}   \\
    
    \Xhline{1.2pt}
    \multirow{9}{*}{AP} 
    &\multirow{3}{*}{\emph{Ohsumed}}  &100  &{0.468} &{0.535}  &{0.423} &{0.522} &{0.549} &\textbf{{0.560}} \\
    ~&~&1,000   &{0.728} &{0.717}  &{0.508} &{0.730} &{0.733} &\textbf{{0.744}}   \\
    ~&~&10,000   &\textbf{{0.830}} &{0.826}  &{0.555} &{0.827} &{0.828} &\textbf{{0.830}}   \\
    \cline{2-9}
    ~&\multirow{3}{*}{\emph{AAPD}}    &200   &{0.566} &{0.624}  &{0.441} &{0.624} &{0.635} &\textbf{{0.639}} \\
    ~&~&2,000    &{0.717} &{0.734}  &{0.505} &\textbf{{0.746}} &{0.732} &{0.740}   \\
    ~&~&20,000   &{0.779} &{0.792}  &{0.539} &{0.784} &{0.777} &\textbf{{0.793}}   \\
    \cline{2-9}
    ~&\multirow{3}{*}{\emph{RCV1-V2}}       &100   &{0.615} &{0.748}  &{0.583} &{0.718} &{0.739} &\textbf{{0.752}} \\
    ~&~&1,000    &{0.876} &{0.884}  &{0.696} &{0.884} &{0.881} &\textbf{{0.888}}   \\
    ~&~&10,000   &{0.926} &{0.924}  &{0.754} &{0.927} &{0.913} &\textbf{{0.933}}   \\
    
    \Xhline{1pt}
    \multicolumn{3}{c||}{Average Rank}  &{3.9} &{3.0} &{5.9}  &{2.4} &{3.2} &\textbf{{1.6}} \\
    \Xhline{1.5pt}
  \end{tabular}
  }
\end{table*}

Generally speaking, our proposed \baby consistently outperforms the baselines. Across all datasets and evaluation metrics, \babyf and \babys rank \emph{1.0} and \emph{3.4} on average, respectively. Several observations are made below: 
(1) First, \baby consistently dominates the pre-training methods (including BERT+CE and VAMPIRE) on both Micro-F1 and Macro-F1 scores by a big margin, especially when labeled texts are scarce. For example, when $N_l=100$, the Macro-F1 scores of \baby are even about 0.18, 0.26 and 0.33 higher than VAMPIRE on the datasets of \emph{AG News}, \emph{Yelp} and \emph{Yahoo}, respectively. Second, when labeled texts are very scarce (\ie when $N_l=100$), \baby performs better than other self-training baseline methods (\ie NB+EM, BERT+AM, VAT, UDA and FreeMatch) on all datasets, \eg for Micro-F1 about 0.14 higher than VAT on \emph{Yahoo}. Otherwise, when labeled texts are large, \baby can also achieve competitive performance, even perform better across all datasets.
(2) Our \baby method consistently outperforms BERT+AM and BERT+CE across all datasets and metrics. For example, when $N_l=100$ the Micro-F1 scores of \baby beat those of BERT+AM by $0.01\sim0.03$ and those of BERT+CE by $0.03\sim0.05$ across all datasets. That is because \baby employs both labeled and unlabeled texts for training and can predict more accurate pseudo-labels of unlabeled texts than BERT+AM, benefiting for the classifier training. This result is expected since \baby performs a Gaussian linear transformation to balance the label angel variances, so as to eliminate the margin bias, leading to more accurate predicted pseudo-labels of unlabeled texts. Besides, these results empirically prove that unlabeled texts are beneficial to the classification performance.
(3) All BERT based methods (\ie BERT+CE, BERT+AM, VAT, UDA, FreeMatch and \baby) consistently dominate baselines based on small models (\ie NB+EM, VAMPIRE). For example, when $N_l=10000$, the Micro-F1 and Macro-F1 scores of BERT+CE are about 0.03, 0.18 and 0.05 higher than those of NB+EM on the datasets of \emph{AG News}, \emph{Yelp} and \emph{Yahoo}, respectively. The observation is expected because BERT is a bigger model, hence can extract more discriminative representations of texts than those from the VAE model used in VAMPIRE and tf-idfs used in NB+EM.

\begin{table*}[htp]
  \centering
  \caption{Experimental results of Micro-F1 and Macro-F1 by varying the number of unlabeled texts $N_u$ on multi-label benchmarks. The best results are highlighted in boldface.}
  \renewcommand\arraystretch{1.0}
  \label{table5-3-1}

\resizebox{0.65\linewidth}{!}{
  \begin{tabular}{c|c|c||p{40pt}<{\centering}p{40pt}<{\centering}p{40pt}<{\centering}p{40pt}<{\centering}p{40pt}<{\centering}}
    \Xhline{1.5pt}
    Metric &Dataset  &$N_u$  &\textbf{BERT+AM} &\textbf{SDRL} &\textbf{CAP} &\textbf{MetaExpert} &\textbf{\baby} \\
    \Xhline{0.5pt}
    \hline
    \hline
    \Xhline{0.5pt}
    \multirow{12}{*}{Micro-F1} 
    &\multirow{4}{*}{\emph{Ohsumed}}  &0 &{0.415}  &{0.308} &{0.373} &{0.389} &\textbf{{0.433}} \\
    ~&~&100   &{0.414}  &{0.316} &{0.370} &{0.350} &\textbf{{0.424}} \\
    ~&~&1,000    &{0.415} &{0.310} &{0.378} &{0.371} &\textbf{{0.435}}   \\
    ~&~&10,000    &{0.421}  &{0.311} &{0.380} &{0.418} &\textbf{{0.447}}   \\
    \cline{2-8}
    ~&\multirow{4}{*}{\emph{AAPD}}    &0 &{0.531}  &{0.372} &{0.486} &{0.491} &\textbf{{0.535}} \\
    ~&~&200  &{0.534}  &{0.371} &{0.488} &{0.505} &\textbf{{0.539}} \\
    ~&~&2,000     &{0.539}  &{0.372} &{0.492} &{0.518} &\textbf{{0.542}}   \\
    ~&~&20,000   &{0.543}  &{0.361} &{0.497} &{0.525} &\textbf{{0.554}}   \\
    \cline{2-8}
    ~&\multirow{4}{*}{\emph{RCV1-V2}}       &0 &{0.641}  &{0.505} &{0.489} &{0.532} &\textbf{{0.643}}  \\
    ~&~&100    &{0.644}  &{0.503} &{0.565} &{0.554} &\textbf{{0.645}} \\
    ~&~&1,000     &{0.646}  &{0.505} &{0.564} &{0.541} &\textbf{{0.652}}   \\
    ~&~&10,000    &{0.649}  &{0.479} &{0.568} &{0.559} &\textbf{{0.655}}   \\
    
    \Xhline{1.2pt}
    \multirow{12}{*}{Macro-F1} 
    &\multirow{4}{*}{\emph{Ohsumed}}  &0 &{0.265}  &{0.077} &{0.245} &\textbf{0.298} &{0.266} \\
    ~&~&100   &{0.258}  &{0.087} &{0.257} &\textbf{0.283} &{0.266} \\
    ~&~&1,000    &{0.275}  &{0.124} &{0.253} &{0.282} &\textbf{{0.286}}   \\
    ~&~&10,000    &{0.298}  &{0.107} &{0.252} &{0.300} &\textbf{{0.308}}   \\
    \cline{2-8}
    ~&\multirow{4}{*}{\emph{AAPD}}    &0 &{0.282} &{0.126} &{0.276} &{0.281} &\textbf{{0.296}} \\
    ~&~&200  &{0.293}  &{0.100} &{0.290} &{0.301} &\textbf{{0.307}} \\
    ~&~&2,000     &{0.318}  &{0.089} &{0.307} &{0.322} &\textbf{{0.328}}   \\
    ~&~&20,000   &{0.342}  &{0.070} &{0.315} &{0.328} &\textbf{{0.356}}   \\
    \cline{2-8}
    ~&\multirow{4}{*}{\emph{RCV1-V2}}       &0 &{0.256}  &{0.144} &{0.194} &{0.189} &\textbf{{0.258}} \\
    ~&~&100    &{0.250}  &{0.104} &{0.227} &{0.208} &\textbf{{0.260}} \\
    ~&~&1,000     &{0.268}  &{0.126} &{0.223} &{0.206} &\textbf{{0.272}}   \\
    ~&~&10,000    &{0.270}  &{0.082} &{0.224} &{0.235} &\textbf{{0.277}}   \\
    \Xhline{1pt}
    \multicolumn{3}{c||}{Average Rank}  &{2.3} &{5.0} &{3.7} &{3.0} &\textbf{{1.1}} \\
    \Xhline{1.5pt}
  \end{tabular}
  }
\end{table*}

\begin{table*}[!ht]
  \centering
  \caption{Experimental results of the ranking loss and average precision (AP) by varying the number of unlabeled texts $N_u$ on multi-label benchmarks. The best results are highlighted in boldface.}
  \renewcommand\arraystretch{1.0}
  \label{table5-3-2}

\resizebox{0.65\linewidth}{!}{
  \begin{tabular}{c|c|c||p{40pt}<{\centering}p{40pt}<{\centering}p{40pt}<{\centering}p{40pt}<{\centering}p{40pt}<{\centering}}
    \Xhline{1.5pt}
    Metric &Dataset  &$N_u$  &\textbf{BERT+AM}  &\textbf{SDRL} &\textbf{CAP} &\textbf{MetaExpert} &\textbf{\baby} \\
    \Xhline{0.5pt}
    \hline
    \hline
    \Xhline{0.5pt}
    \multirow{12}{*}{Ranking Loss} 
    &\multirow{4}{*}{\emph{Ohsumed}}  &0 &{0.195}  &{0.253} &{0.213} &{0.234} &\textbf{{0.192}} \\
    ~&~&100   &{0.200}  &{0.251} &{0.206} &{0.205} &\textbf{{0.189}} \\
    ~&~&1,000    &{0.198}  &{0.256} &{0.207} &{0.205} &\textbf{{0.187}}   \\
    ~&~&10,000    &{0.208}  &{0.249} &{0.199} &{0.192} &\textbf{{0.184}}   \\
    \cline{2-8}
    ~&\multirow{4}{*}{\emph{AAPD}}    &0 &\textbf{{0.103}}  &{0.193} &{0.112} &{0.116} &{0.104} \\
    ~&~&200  &{0.106}  &{0.190} &{0.110} &{0.107} &\textbf{{0.102}} \\
    ~&~&2,000     &{0.112}  &{0.201} &{0.110} &{0.110} &\textbf{{0.109}}   \\
    ~&~&20,000   &{0.110}  &{0.196} &\textbf{{0.096}} &\textbf{0.096} &{0.105}   \\
    \cline{2-8}
    ~&\multirow{4}{*}{\emph{RCV1-V2}}       &0 &{0.052}  &{0.100} &{0.081} &{0.056} &\textbf{{0.050}} \\
    ~&~&100    &{0.051}  &{0.088} &{0.064} &{0.058} &\textbf{{0.049}} \\
    ~&~&1,000     &{0.056}  &{0.090} &{0.058} &{0.054} &\textbf{{0.053}}   \\
    ~&~&10,000    &{0.047}  &{0.090} &{0.056} &\textbf{0.046} &{0.047}   \\
    \Xhline{1.2pt}
    \multirow{12}{*}{AP} 
    &\multirow{4}{*}{\emph{Ohsumed}}  &0 &{0.528}  &{0.414} &{0.515} &{0.527} &\textbf{{0.544}} \\
    ~&~&100   &{0.525}  &{0.422} &{0.520} &\textbf{0.545} &{0.536} \\
    ~&~&1,000    &{0.528}  &{0.419} &{0.519} &{0.531} &\textbf{{0.545}}   \\
    ~&~&10,000    &{0.535}  &{0.423} &{0.522} &{0.549} &\textbf{{0.560}}   \\
    \cline{2-8}
    ~&\multirow{4}{*}{\emph{AAPD}}    &0 &{0.625} &{0.461} &{0.597} &{0.607} &\textbf{{0.627}} \\
    ~&~&200  &{0.628}  &{0.460} &{0.605} &{0.624} &\textbf{{0.630}} \\
    ~&~&2,000     &\textbf{0.633}  &{0.458} &{0.607} &{0.626} &\textbf{{0.633}}   \\
    ~&~&20,000   &{0.624}  &{0.441} &{0.624} &{0.635} &\textbf{{0.639}}   \\
    \cline{2-8}
    ~&\multirow{4}{*}{\emph{RCV1-V2}}       &0 &{0.739}  &{0.605} &{0.681} &{0.722} &\textbf{{0.742}} \\
    ~&~&100    &{0.746}  &{0.600} &{0.714} &{0.719} &\textbf{{0.747}} \\
    ~&~&1,000     &{0.737}  &{0.607} &{0.719} &{0.737} &\textbf{{0.740}}   \\
    ~&~&10,000    &{0.748}  &{0.583} &{0.718} &{0.739} &\textbf{{0.752}}   \\
    \Xhline{1pt}
    \multicolumn{3}{c||}{Average Rank}  &{2.4} &{5.0} &{3.7} &{2.6} &\textbf{{1.3}} \\
    \Xhline{1.5pt}
  \end{tabular}
  }
\end{table*}

\subsubsection{Varying Number of Unlabeled Texts}
For NB+EM, BERT+AM, VAMPIRE, VAT, UDA, FreeMatch and \baby, we also perform the experiments with 100 labeled texts and varying the number of unlabeled texts $N_u$ over the set $\{0, 200, 2000, 20000\}$ for \emph{AG News} and \emph{Yelp}, and $\{0, 400, 4000, 40000\}$ for \emph{Yahoo}. Note that VAMPIRE needs unlabeled texts for pre-training, thus we omit the experiments for VAMPIRE with $N_u=0$. The classification results are reported in Table \ref{table5-3}. Roughly, for all methods the classification performance becomes better as the amount of unlabeled texts increasing. For instance, the Micro-F1 scores of \baby on all datasets gain about 0.3 improvement as the number of unlabeled texts increasing. These results prove the effectiveness of unlabeled texts in riching the limited supervision from scarce labeled texts and improving the classification performance. Besides, an obvious observation is that the self-training methods (\ie NB+EM, BERT+AM, VAT, UDA and \baby) consistently outperform the pre-training method (\ie VAMPIRE), especially when unlabeled texts are fewer. The possible reason is that the pre-training methods need more unlabeled texts for pre-training while the self-training methods do not have the requirement.

\subsection{Results of Semi-Supervised Multi-Label Classification}
\label{4.3}

\subsubsection{Varying Number of Labeled Texts}

In this case, we conduct the experiments by varying the number of labeled texts $N_l$ over the set $\{100,1000,10000\}$ with the number of unlabeled texts $N_u=10000$ for \emph{Ohsumed} and \emph{RCV1-V2}, and $N_l$ over $\{200,2000,20000\}$ with $N_u=20000$ for \emph{AAPD}. The classification results over all datasets are illustrated in Tables \ref{table5-2-1} and \ref{table5-2-2}, in which the best scores among all comparing baselines are highlighted in boldface. 

Generally speaking, our proposed \baby consistently outperforms the baselines on the classification metrics, and performs better than them on the ranking metrics in most cases. 
Across all datasets, \baby ranks \emph{1.0} and \emph{1.4} on average over both label-based classification metrics and example-based ranking metrics, respectively. 
As shown in Tables \ref{table5-2-1} and \ref{table5-2-2}: 
(1) First, our \baby consistently performs better than all baselines on both Micro-F1 and Macro-F1 scores. For example, the Micro-F1 scores of \baby beat those of current semi-supervised multi-label classification baselines CAP and SDRL by $0.03\sim0.10$ and $0.14\sim0.28$ across all settings, respectively. Second, on the ranking metrics, \ie Ranking Loss and AP, \baby also outperforms all baselines in most cases, such as for AP about 0.04 and 0.14 higher than CAP and SDRL on \emph{Ohsumed}, respectively.
(2) Our \baby method consistently dominates BERT+AM and BERT+BCE across all datasets and metrics when labeled texts are scarce. For example, when $N_l=100$ for \emph{Ohsumed} the Micro-F1 and AP scores of \baby beat those of BERT+AM by 0.026 and 0.025, and those of BERT+BCE by 0.169 and 0.092, respectively. These results are similar to ones in the semi-supervised multi-class text classification case of Section \ref{4.2}, and demonstrate the effectiveness of our proposed BDD loss again. Besides, when labeled texts are large, compared with BERT+AM and BERT+BCE, \baby can also achieve competitive performance, even perform better across all datasets.

\subsubsection{Varying Number of Unlabeled Texts}

For BERT+AM, SDRL, CAP and \baby, we also perform the experiments with 100 labeled texts and varying the number of unlabeled texts $N_u$ over the set $\{0, 100, 1000, 10000\}$ for \emph{Ohsumed} and \emph{RCV1-V2}, and $N_l=200$ and varying $N_u$ over $\{0, 200, 2000, 20000\}$ for \emph{AAPD}. The results are shown in Tables \ref{table5-3-1} and \ref{table5-3-2}. 
Generally, all methods perform better as the amount of unlabeled texts increasing across all benchmarks and metrics. 
For instance, as the number of unlabeled texts rising, the Macro-F1 scores of \baby accumulate about $0.02\sim 0.06$ improvement on \emph{Ohsumed}, \emph{AAPD} and \emph{RCV1-V2}. These results demonstrate the effectiveness of unlabeled texts in riching the limited supervision from scarce labeled texts and improving the classification performance, especially for the rare categories in the multi-label case.
Besides, an obvious observation is that the thresholding methods (\ie BERT+AM, CAP and \baby) consistently outperform SDRL. The possible reason is that SDRL introduces more noisy data because it continuously adds pseudo-labeled unlabeled texts into the labeled text set without discarding.

\begin{table*}[htp]
  \centering
  \caption{Classification performance on the multi-class text classification benchmark \emph{AG News} with 100 labeled data and 20,000 unlabeled data, and the multi-label text classification benchmark \emph{Ohsumed} with 100 labeled data and 10,000 unlabeled data after removing different parts of \baby.}
  \renewcommand\arraystretch{1.1}
  \label{table5-4}
      \begin{tabular}{p{70pt}<{\centering}p{35pt}<{\centering}p{35pt}<{\centering}|p{35pt}<{\centering}p{35pt}<{\centering}|p{35pt}<{\centering}p{35pt}<{\centering}p{45pt}<{\centering}p{35pt}<{\centering}}
        \Xhline{1.5pt}
        \multirow{2}{*}{Model}     &\multicolumn{2}{c|}{Multi-class (\textsc{-s})} &\multicolumn{2}{c|}{Multi-class (\textsc{-f})}  &\multicolumn{4}{c}{Multi-label}  \\
        \cline{2-9}
        ~ &Micro-F1  &Macro-F1 &Micro-F1  &Macro-F1  &Micro-F1 &Macro-F1 &Ranking Loss &AP\\
        \Xhline{1pt}
        \textbf{\baby}                  &\textbf{0.872} &\textbf{0.872} &\textbf{0.876} &\textbf{0.876} &\textbf{0.447} &\textbf{0.308} &\textbf{0.184} &\textbf{0.560} \\
        \Xhline{0.5pt}
        -regularization  &{0.863} &{0.864} &{0.874} &{0.873} &{0.428} &{0.305} &{0.195} &{0.544} \\
        -BDD                   &{0.856} &{0.856} &{0.870} &{0.870} &{0.421} &{0.298} &{0.208} &{0.535} \\
        -unlabeled texts        &{0.844} &{0.843} &{0.750} &{0.751} &{0.433} &{0.266} &{0.192} &{0.544} \\
        -all                   &{0.839} &{0.840} &{0.839} &{0.840} &{0.384} &{0.222} &{0.220} &{0.494} \\
        \Xhline{1.5pt}
      \end{tabular}
\end{table*}

\subsection{Ablation Study} 
We perform ablation studies by stripping each component each time to examine the effectiveness of each component in \baby for both semi-supervised multi-class and semi-supervised multi-label text classification scenarios. Here, we denote ``Multi-class (\textsc{-s})'' and ``Multi-class (\textsc{-f})'' as the cases of \babys and \babyf, respectively; ``BDD'' as balanced deep representation angular loss $\mathfrak{L}_{bdd}$ in Eq.\ref{eq3-1}; and ``regularization'' as the entropy regularization term of Eq.\eqref{eq3-4} in the multi-class case and the low-rank regularization term of Eq.\eqref{eq3-7} in the multi-label case. 
Stripping BDD means that we replace the proposed loss $\mathfrak{L}_{bdd}$ with the AM loss $\mathfrak{L}_{am}$ in Eq.\ref{eq3-1-1}. 
And Stripping regularization means that we set $\lambda_2=0$ and $\lambda_3=0$.
The results are displayed in Table \ref{table5-4}. Overall, the classification performance will drop when removing any component of \baby, suggesting that all parts make contributions to the final performance of \baby. Besides, removing unlabeled texts brings the most significant drop of the performance. This result is expected because label angle variances approximated only with very scarce labeled texts will have lower accuracy, resulting in worse performance. Further, in contrast to regularization, the performance after stripping BDD decreases more. Note that the difference between the proposed $\mathfrak{L}_{bdd}$ and $\mathfrak{L}_{am}$ is whether constraining the label angle variances to be balanced or not. This result indicates that the balanced constraint of label angle variances brings a better deep classifier as well as more accurate pseudo-labels for unlabeled texts, especially when labeled texts are limited, and also empirically proves the effectiveness of our balanced label angle variances.

\begin{figure}
\includegraphics[width=0.45\textwidth]{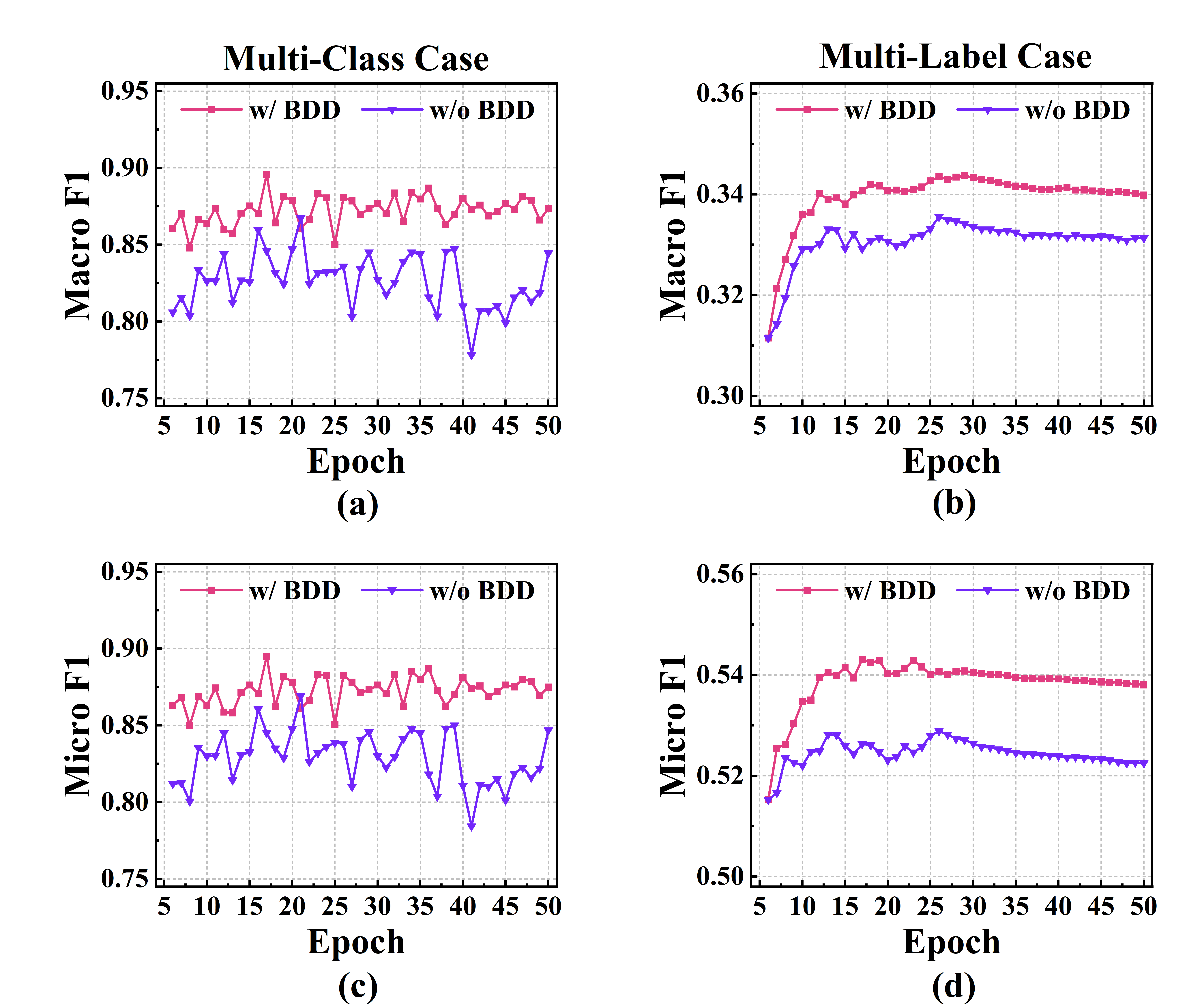}
\centering
\caption{The accuracy of pseudo-labels during the training procedure with or without BDD loss (w/ BDD and w/o BDD) across \emph{AG News} (\textbf{Multi-Class Case}, $N_l=100, N_u=20000$) and \emph{AAPD} (\textbf{Multi-Label Case}, $N_l=200, N_u=20000$), respectively.}
\label{Fig4}
\end{figure}

\subsection{Accuracy of Pseudo-Labeling}

To evaluate the effectiveness of balancing the label angle variances in improving the pseudo-labeling accuracy, we calculate the accuracy of pseudo-labels of unlabeled texts during the training procedure of BERT+AM and \baby (\babyf for multi-class case), \ie with or without BDD loss (w/ BBD and w/o BDD), across \emph{AG News} with $N_l=100, N_u=20000$ and \emph{AAPD} with $N_l=200, N_u=20000$, respectively. The corresponding Macro F1 and Micro F1 scores are shown in Figure \ref{Fig4}. Overall, the pseudo-labeling accuracy with BDD loss consistently outperforms that without BDD loss on both multi-class and multi-label cases. This result is expected since \baby employs the BDD loss to balance the label angel variances, so as to eliminate the margin bias, leading to more accurate predicted pseudo-labels of unlabeled texts.

%% file: Sec_Discussion.tex
\section{Discussion and Implications}

\subsection{Discussion}
According to the experimental results (as shown in Tables \ref{table5-2}--\ref{table5-3-2}), our proposed \baby can achieve better performance than the SSTC baseline methods in both multi-class and multi-label cases over different sizes of labeled and unlabeled datasets. As shown in the ablative results (Table \ref{table5-4}), one can find that each part of \baby contributes to its final performance, and the BDD loss makes a much more significant contribution to performance improvement than the regularization term does.


We also monitor the accuracy of pseudo-labels assigned to unlabeled texts during model tuning with or without BDD loss on both multi-class and multi-label cases in Fig.\ref{Fig4}. It can be observed that the accuracy of pseudo-labels with BDD loss is much higher than that without BDD loss on both multi-class and multi-label cases. This indicates that our proposed BDD loss can improve the accuracy of pseudo-labels for unlabeled texts effectively during model tuning as it balances the label angle variance, thereby eliminating the margin bias.



\subsection{Implications}
In this paper, we argue that due to the margin bias problem caused by the large difference between representation distributions of labels
in SSTC, current methods working in the self-training spirit suffer from the low accuracy of pseudo-labels. We verify the margin bias problem by evaluating the average difference in label angle variances computed in semi-supervised and supervised manners for both multi-class and multi-label cases, and illustrate the corresponding effect with a toy example. To constrain the label angle variances to be balanced, we design a novel BDD loss by incorporating several simple Gaussian linear transformations into the AM loss, and propose a novel self-training SSTC method \baby for both multi-class and multi-label manners.

This study identifies a potential reason, \ie margin bias problem, causing the low accuracy of pseudo-labels within unlabeled texts in current self-training SSTC methods, encompassing both multi-class and multi-label manners. To address the margin bias, it introduces a simple yet effective mechanism by balancing representation distributions of labels with several linear transformations based on the Gaussian distribution assumption, and implements both multi-class and multi-label versions. The experimental results demonstrate that balancing representation distributions can enhance the accuracy of pseudo-labels during the self-training procedure, resulting in improved SSTC performance. In practice, there are many complex scenarios, such as multi-model distributions for each class (\eg some label may have a naturally high variance in examples). The truncated Gaussian distribution, mixture-of-Gaussian distribution, or the widely-used von Mises-Fisher distribution could handle these cases, and we will exploit them in future work.

%% file: Sec_Conclusion.tex
\section{Conclusion and Future Works}
\label{5}
In this paper, we propose a novel self-training SSTC method, namely \baby. Our \baby addresses the margin bias problem in SSTC by balancing the label angle variances, \ie the variance of label angles of texts within the same label. We estimate the label angle variances with both labeled and unlabeled texts during the self-training loops. To constrain the label angle variances to be balanced, we design several Gaussian linear transformations and incorporate them into a well-established AM loss. We implement both multi-class classification and multi-label classification versions of \baby by introducing some pseudo-labeling tricks and regularization terms. Our \baby empirically outperforms the existing SSTC baseline methods in both semi-supervised multi-class and semi-supervised multi-label cases.


Although the proposed \baby can achieve competitive SSTC performance in both semi-supervised multi-class and semi-supervised multi-label cases, it exhibits several limitations under specific scenarios, such as imbalanced and long-tailed semi-supervised text classification, open-set/open-world semi-supervised learning where unlabeled data includes some out-of-distribution samples, and noisy labeled data. First, balancing representation distributions for imbalanced, long-tailed classes, or ones with multi-model distributions may cause an underestimate or overestimate of their variances. Second, out-of-distribution unlabeled samples and noisy labeled samples will decrease the approximate accuracy of representation distributions and introduce additional noise into pseudo-labels.

%% file: SSTC.bib
@inproceedings{Jialun2020,
  author    = "Jialun Liu and others",
  title     = "Deep Representation Learning on Long-tailed Data: A Learnable Embedding Augmentation Perspective",
  booktitle = "IEEE CVPR",
  publisher = "",
  pages     = "2970--2979",
  address   = "",
  month     = "",
  year      = "2020"
}

@article{SSL2020,
  author    = "Jesper E. van Engelen and Holger H. Hoos",
  title     = "A survey on semi-supervised learning",
  journal   = "Machine Learning",
  volume    = "109",
  number    = "2",
  pages     = "373--440",
  month     = "",
  year      = "2020"
}

@inproceedings{MixText2020,
  author    = "Jiaao Chen and Zichao Yang and Diyi Yang",
  title     = "MixText: Linguistically-Informed Interpolation of Hidden Space for Semi-Supervised Text Classification",
  booktitle = "ACL",
  publisher = "",
  pages     = "2147--2157",
  address   = "",
  month     = "",
  year      = "2020"
}

@inproceedings{CVT2018,
  author    = "Kevin Clark and others",
  title     = "Semi-Supervised Sequence Modeling with Cross-View Training",
  booktitle = "EMNLP",
  publisher = "",
  pages     = "1914--1925",
  address   = "",
  month     = "",
  year      = "2018"
}

@inproceedings{LSTMMIXED2019,
  author    = "Devendra Singh Sachan and Manzil Zaheer and Ruslan Salakhutdinov",
  title     = "Revisiting {LSTM} Networks for Semi-Supervised Text Classification via Mixed Objective Function",
  booktitle = "AAAI",
  publisher = "",
  pages     = "6940--6948",
  address   = "",
  month     = "",
  year      = "2019"
}

@inproceedings{VAM2019,
  author    = "Suchin Gururangan and others",
  title     = "Variational Pretraining for Semi-supervised Text Classification",
  booktitle = "ACL",
  publisher = "",
  pages     = "5880--5894",
  address   = "",
  month     = "",
  year      = "2019"
}

@inproceedings{BERT2019,
  author    = "Jacob Devlin and others",
  title     = "{BERT:} Pre-training of Deep Bidirectional Transformers for Language Understanding",
  booktitle = "NAACL",
  publisher = "",
  pages     = "4171--4186",
  address   = "",
  month     = "",
  year      = "2019"
}

@inproceedings{UDA2020,
  title     = "Unsupervised Data Augmentation for Consistency Training",
  author    = "Qizhe Xie and others",
  booktitle = "NeurIPS",
  publisher = "",
  pages     = "",
  address   = "",
  month     = "",
  year      = "2020"
}

@article{VAT2019,
  author    = "Takeru Miyato and others",
  title     = "Virtual Adversarial Training: {A} Regularization Method for Supervised and Semi-Supervised Learning",
  journal   = "{IEEE} TPAMI",
  volume    = "41",
  number    = "8",
  pages     = "1979--1993",
  month     = "",
  year      = "2019"
}

@inproceedings{VAT2017,
  author    = "Takeru Miyato and Andrew M. Dai and Ian J. Goodfellow",
  title     = "Adversarial Training Methods for Semi-Supervised Text Classification",
  booktitle = "ICLR",
  publisher = "",
  pages     = "",
  address   = "",
  month     = "",
  year      = "2017"
}

@inproceedings{Data2015,
  author    = "Xiang Zhang and Junbo Jake Zhao and Yann LeCun",
  title     = "Character-level Convolutional Networks for Text Classification",
  booktitle = "NeurIPS",
  publisher = "",
  pages     = "649--657",
  address   = "",
  month     = "",
  year      = "2015"
}

@inproceedings{Yahoo2008,
  author    = "Ming{-}Wei Chang and others",
  title     = "Importance of Semantic Representation: Dataless Classification",
  booktitle = "AAAI",
  publisher = "",
  pages     = "830--835",
  address   = "",
  month     = "",
  year      = "2008"
}

@article{GPT32020,
  author    ="Tom B. Brown and others",
  title     ="Language Models are Few-Shot Learners",
  journal   ="arXiv preprint arXiv:2005.14165",
  volume    ="",
  number    = "",
  pages     = "",
  month     = "",
  year      ="2020"
}

@inproceedings{ULMFit2018,
  author    ="Jeremy Howard and Sebastian Ruder",
  title     ="Universal Language Model Fine-tuning for Text Classification",
  booktitle ="ACL",
  publisher ="",
  pages     ="328--339",
  address   = "",
  month     = "",
  year      ="2018"
}

@inproceedings{ELMO2018,
  author    = "Matthew E. Peters and others",
  title     = "Deep Contextualized Word Representations",
  booktitle = "NAACL",
  publisher = "",
  pages     = "2227--2237",
  address   = "",
  month     = "",
  year      = "2018"
}

@inproceedings{XLNet2019,
  author    ="Zhilin Yang and others",
  title     ="XLNet: Generalized Autoregressive Pretraining for Language Understanding",
  booktitle ="NeurIPS",
  publisher = "",
  pages     ="5753--5763",
  address   = "",
  month     = "",
  year      ="2019"
}

@article{GPT22019,
  author    = "Alec Radford and others",
  title     = "Language Models are Unsupervised Multitask Learners",
  journal   = "",
  volume    = "",
  number    = "",
  pages     = "",
  month     = "",
  year      = "2019"
}

@article{GPT2018,
  author    = "Alec Radford and others",
  title     = "Improving Language Understanding by Generative Pre-Training",
  journal   = "",
  volume    = "",
  number    = "",
  pages     = "",
  month     = "",
  year      = "2019"
}

@inproceedings{Dai2015,
  author    = "Andrew M. Dai and Quoc V. Le",
  title     = "Semi-supervised Sequence Learning",
  booktitle = "NeurIPS",
  publisher = "",
  pages     = "3079--3087",
  address   = "",
  month     = "",
  year      = "2015"
}

@inproceedings{EntMin2004,
  author    = "Yves Grandvalet and Yoshua Bengio",
  title     = "Semi-supervised Learning by Entropy Minimization",
  booktitle = "NeurIPS",
  publisher = "",
  pages     = "529--536",
  address   = "",
  month     = "",
  year      = "2004"
}

@inproceedings{RSTP2016,
  author    = "Mehdi Sajjadi and Mehran Javanmardi and Tolga Tasdizen",
  title     = "Regularization With Stochastic Transformations and Perturbations for Deep Semi-Supervised Learning",
  booktitle = "NeurIPS",
  publisher = "",
  pages     = "1171--1179",
  address   = "",
  month     = "",
  year      = "2016"
}

@inproceedings{CosFace2018,
  author    = "Hao Wang and others",
  title     = "CosFace: Large Margin Cosine Loss for Deep Face Recognition",
  booktitle = "{IEEE} CVPR",
  publisher = "",
  pages     = "5265--5274",
  address   = "",
  month     = "",
  year      = "2018"
}

@article{NBMEM2000,
  author    = "Kamal Nigam and others",
  title     = "Text Classification from Labeled and Unlabeled Documents using {EM}",
  journal   = "Machine Learning",
  volume    = "39",
  number    = "2",
  pages     = "103--134",
  month     = "",
  year      = "2000"
}

@article{SKLEARN2011,
  author    = "Fabian Pedregosa and others",
  title     = "Scikit-learn: Machine Learning in Python",
  journal   = "JMLR",
  volume    = "12",
  number    = "",
  pages     = "2825--2830",
  month     = "",
  year      = "2011"
}

@inproceedings{CAP2023,
  author    = "Ming{-}Kun Xie and others",
  title     = "Class-Distribution-Aware Pseudo Labeling for Semi-Supervised Multi-Label Learning",
  booktitle = "NeurIPS",
  publisher = "",
  pages     = "",
  address   = "",
  month     = "",
  year      = "2023"
}

@article{ADMM2011,
  author    = "Stephen P. Boyd and others",
  title     = "Distributed Optimization and Statistical Learning via the Alternating Direction Method of Multipliers",
  journal   = "Foundations and Trends in Machine Learning",
  volume    = "3",
  number    = "1",
  pages     = "1--122",
  month     = "",
  year      = "2011"
}

@article{SVTA2010,
  author    = "Jian{-}Feng Cai and Emmanuel J. Cand{\`{e}}s and Zuowei Shen",
  title     = "A Singular Value Thresholding Algorithm for Matrix Completion",
  journal   = "{SIAM} Journal on Optimization",
  volume    = "20",
  number    = "4",
  pages     = "1956--1982",
  month     = "",
  year      = "2010"
}

@inproceedings{BDD2021,
  author    = "Changchun Li and Ximing Li and Jihong Ouyang",
  title     = "Semi-Supervised Text Classification with Balanced Deep Representation Distributions",
  booktitle = "ACL",
  publisher = "",
  pages     = "5044--5053",
  address   = "",
  month     = "",
  year      = "2021"
}

@article{DRML2021,
  author    = "Lichen Wang and others",
  title     = "Semi-Supervised Dual Relation Learning for Multi-Label Classification",
  journal   = "{IEEE} Transactions on Image Processing",
  volume    = "30",
  number    = "",
  pages     = "9125--9135",
  month     = "",
  year      = "2021"
}

@inproceedings{DRML2020,
  author    = "Lichen Wang and others",
  title     = "Dual Relation Semi-Supervised Multi-Label Learning",
  booktitle = "AAAI",
  publisher = "",
  pages     = "6227--6234",
  address   = "",
  month     = "",
  year      = "2020"
}

@inproceedings{SoftMatch2023,
  author    = "Hao Chen and others",
  title     = "SoftMatch: Addressing the Quantity-Quality Trade-off in Semi-supervised Learning",
  booktitle = "ICLR",
  publisher = "",
  pages     = "",
  address   = "",
  month     = "",
  year      = "2023"
}

@inproceedings{FreeMatch2023,
  author    = "Yidong Wang and others",
  title     = "FreeMatch: Self-adaptive Thresholding for Semi-supervised Learning",
  booktitle = "ICLR",
  publisher = "",
  pages     = "",
  address   = "",
  month     = "",
  year      = "2023"
}

@inproceedings{MetaExpert2025,
  author    = "Yaxin Hou and Yuheng Jia",
  title     = "A Square Peg in a Square Hole: Meta-Expert for Long-Tailed Semi-Supervised Learning",
  booktitle = "ICML",
  publisher = "",
  pages     = "",
  address   = "",
  month     = "",
  year      = "2025"
}

@inproceedings{ALMLL2021,
  author       = "Tal Ridnik and others",
  title        = "Asymmetric Loss For Multi-Label Classification",
  booktitle    = "{IEEE} ICCV",
  pages        = "82--91",
  publisher    = "",
  address      = "",
  month        = "",
  year         = "2021"
}

@inproceedings{TWMLL2023,
  author       = "Takumi Kobayashi",
  title        = "Two-Way Multi-Label Loss",
  booktitle    = "{IEEE/CVF} CVPR",
  pages        = "7476--7485",
  publisher    = "",
  address      = "",
  month        = "",
  year         = "2023"
}

@inproceedings{SSMLLDSGM2020,
  author       = "Wanli Shi and others",
  title        = "Semi-Supervised Multi-Label Learning from Crowds via Deep Sequential Generative Model",
  booktitle    = "{ACM} {SIGKDD}",
  pages        = "1141--1149",
  publisher    = "",
  address      = "",
  month        = "",
  year         = "2020"
}

@inproceedings{SSMLLGD2021,
  author       = "Zixing Song and others",
  title        = "Semi-Supervised Multi-label Learning for Graph-Structured Data",
  booktitle    = "{ACM} CIKM",
  pages        = "1723--1733",
  publisher    = "",
  address      = "",
  month        = "",
  year         = "2021"
}

@inproceedings{LSDR2019,
  author       = "Lin Xiao and others",
  title        = "Label-Specific Document Representation for Multi-Label Text Classification",
  booktitle    = "EMNLP",
  pages        = "466--475",
  publisher    = "",
  address      = "",
  month        = "",
  year         = "2019"
}

@inproceedings{CLENNM2022,
  author       = "Xi'ao Su and Ran Wang and Xinyu Dai",
  title        = "Contrastive Learning-Enhanced Nearest Neighbor Mechanism for Multi-Label Text Classification",
  booktitle    = "ACL",
  pages        = "672--679",
  publisher    = "",
  address      = "",
  month        = "",
  year         = "2022"
}

@inproceedings{BILLC2016,
  author       = "Gakuto Kurata and Bing Xiang and Bowen Zhou",
  title        = "Improved Neural Network-based Multi-label Classification with Better Initialization Leveraging Label Co-occurrence",
  booktitle    = "NAACL",
  pages        = "521--526",
  publisher    = "",
  address      = "",
  month        = "",
  year         = "2016"
}

@inproceedings{CNNSC2014,
  author       = "Yoon Kim",
  title        = "Convolutional Neural Networks for Sentence Classification",
  booktitle    = "EMNLP",
  pages        = "1746--1751",
  publisher    = "",
  address      = "",
  month        = "",
  year         = "2014"
}

@inproceedings{RNNTC2016,
  author       = "Pengfei Liu and Xipeng Qiu and Xuanjing Huang",
  title        = "Recurrent Neural Network for Text Classification with Multi-Task Learning",
  booktitle    = "IJCAI",
  pages        = "2873--2879",
  publisher    = "",
  address      = "",
  month        = "",
  year         = "2016"
}

@inproceedings{SGM2018,
  author       = "Pengcheng Yang and others",
  title        = "{SGM:} Sequence Generation Model for Multi-label Classification",
  booktitle    = "International Conference on Computational Linguistics",
  pages        = "3915--3926",
  publisher    = "",
  address      = "",
  month        = "",
  year         = "2018"
}

@inproceedings{LSDGNN2021,
  author       = "Qianwen Ma and others",
  title        = "Label-Specific Dual Graph Neural Network for Multi-Label Text Classification",
  booktitle    = "ACL",
  pages        = "3855--3864",
  publisher    = "",
  address      = "",
  month        = "",
  year         = "2021"
}

@inproceedings{GMVACL2022,
  author       = "Junwen Bai and Shufeng Kong and Carla P. Gomes",
  title        = "Gaussian Mixture Variational Autoencoder with Contrastive Learning for Multi-Label Classification",
  booktitle    = "ICML",
  pages        = "1383--1398",
  publisher    = "",
  address      = "",
  month        = "",
  year         = "2022"
}

@inproceedings{TPT2020,
  author       = "Wei{-}Cheng Chang and others",
  title        = "Taming Pretrained Transformers for Extreme Multi-label Text Classification",
  booktitle    = "{ACM} {SIGKDD} Conference on Knowledge Discovery and Data Mining",
  pages        = "3163--3171",
  publisher    = "",
  address      = "",
  month        = "",
  year         = "2020"
}

@inproceedings{BMMLTC2021,
  author       = "Yi Huang and others",
  title        = "Balancing Methods for Multi-Label Text Classification with Long-Tailed Class Distribution",
  booktitle    = "EMNLP",
  pages        = "8153--8161",
  publisher    = "",
  address      = "",
  month        = "",
  year         = "2021"
}

@inproceedings{DLEMLTC2017,
  author       = "Jingzhou Liu and others",
  title        = "Deep Learning for Extreme Multi-Label Text Classification",
  booktitle    = "{ACM} {SIGIR} International Conference on Research and Development in Information Retrieval",
  pages        = "115--124",
  publisher    = "",
  address      = "",
  month        = "",
  year         = "2017"
}

@inproceedings{AttentionXML2019,
  author       = "Ronghui You and others",
  title        = "AttentionXML: Label Tree-based Attention-Aware Deep Model for High-Performance Extreme Multi-Label Text Classification",
  booktitle    = "NeurIPS",
  pages        = "5812--5822",
  publisher    = "",
  address      = "",
  month        = "",
  year         = "2019"
}

@article{BSF2022,
  author       = "Min{-}Ling Zhang and Jun{-}Peng Fang and Yi{-}Bo Wang",
  title        = "BiLabel-Specific Features for Multi-Label Classification",
  journal      = "{ACM} Transactions on Knowledge Discovery from Data",
  volume       = "16",
  number       = "1",
  pages        = "18:1--18:23",
  month        = "",
  year         = "2022"
}

@inproceedings{CGP2021,
  author       = "Qian{-}Wen Zhang and others",
  title        = "Correlation-Guided Representation for Multi-Label Text Classification",
  booktitle    = "IJCAI",
  pages        = "3363--3369",
  publisher    = "",
  address      = "",
  month        = "",
  year         = "2021"
}

@inproceedings{DEMLL2018,
  author       = "Wenjie Zhang and others",
  title        = "Deep Extreme Multi-Label Learning",
  booktitle    = "{ACM} International Conference on Multimedia Retrieval",
  pages        = "100--107",
  publisher    = "",
  address      = "",
  month        = "",
  year         = "2018"
}

@inproceedings{FixMatch2020,
  author       = "Kihyuk Sohn and others",
  title        = "FixMatch: Simplifying Semi-Supervised Learning with Consistency and Confidence",
  booktitle    = "NeurIPS",
  pages        = "",
  publisher    = "",
  address      = "",
  month        = "",
  year         = "2020"
}

@inproceedings{FlexMatch2021,
  author       = "Bowen Zhang and others",
  title        = "FlexMatch: Boosting Semi-Supervised Learning with Curriculum Pseudo Labeling",
  booktitle    = "NeurIPS",
  pages        = "18408--18419",
  publisher    = "",
  address      = "",
  month        = "",
  year         = "2021"
}

@inproceedings{PGPL2023,
  author       = "Weiyi Yang and others",
  title        = "Prototype-Guided Pseudo Labeling for Semi-Supervised Text Classification",
  booktitle    = "ACL",
  pages        = "16369--16382",
  publisher    = "",
  address      = "",
  month        = "",
  year         = "2023"
}

@inproceedings{PCSM2022,
  author       = "Hai{-}Ming Xu and Lingqiao Liu and Ehsan Abbasnejad",
  title        = "Progressive Class Semantic Matching for Semi-supervised Text Classification",
  booktitle    = "NAACL",
  pages        = "3003--3013",
  publisher    = "",
  address      = "",
  month        = "",
  year         = "2022"
}

@inproceedings{SALNet2021,
  author       = "Ju Hyoung Lee and Sang{-}Ki Ko and Yo{-}Sub Han",
  title        = "SALNet: Semi-supervised Few-Shot Text Classification with Attention-based Lexicon Construction",
  booktitle    = "AAAI",
  pages        = "13189--13197",
  publisher    = "",
  address      = "",
  month        = "",
  year         = "2021"
}

@inproceedings{Dash2021,
  author       = "Yi Xu and others",
  title        = "Dash: Semi-Supervised Learning with Dynamic Thresholding",
  booktitle    = "ICML",
  pages        = "11525--11536",
  publisher    = "",
  address      = "",
  month        = "",
  year         = "2021"
}

@inproceedings{FMRTF2021,
  author       = "Jiong Zhang and others",
  title        = "Fast Multi-Resolution Transformer Fine-tuning for Extreme Multi-label Text Classification",
  booktitle    = "NeurIPS",
  pages        = "7267--7280",
  publisher    = "",
  address      = "",
  month        = "",
  year         = "2021"
}

@inproceedings{LightXML2021,
  author       = "Ting Jiang and others",
  title        = "LightXML: Transformer with Dynamic Negative Sampling for High-Performance Extreme Multi-label Text Classification",
  booktitle    = "AAAI",
  pages        = "7987--7994",
  publisher    = "",
  address      = "",
  month        = "",
  year         = "2021"
}

@inproceedings{DVAE2020,
  author       = "Junwen Bai and Shufeng Kong and Carla P. Gomes",
  title        = "Disentangled Variational Autoencoder based Multi-Label Classification with Covariance-Aware Multivariate Probit Model",
  booktitle    = "IJCAI",
  pages        = "4313--4321",
  publisher    = "",
  address      = "",
  month        = "",
  year         = "2020"
}

@inproceedings{HOTVAE2021,
  author       = "Wenting Zhao and others",
  title        = "{HOT-VAE:} Learning High-Order Label Correlation for Multi-Label Classification via Attention-Based Variational Autoencoders",
  booktitle    = "AAAI",
  pages        = "15016--15024",
  publisher    = "",
  address      = "",
  month        = "",
  year         = "2021"
}

@article{CLLS2022,
  author       = "Jun{-}Yi Hang and Min{-}Ling Zhang",
  title        = "Collaborative Learning of Label Semantics and Deep Label-Specific Features for Multi-Label Classification",
  journal      = "{IEEE} TPAMI",
  volume       = "44",
  number       = "12",
  pages        = "9860--9871",
  month        = "",
  year         = "2022"
}

@article{RCV12004,
  author       = "David D. Lewis and others",
  title        = "{RCV1:} {A} New Benchmark Collection for Text Categorization Research",
  journal      = "JMLR",
  volume       = "5",
  number       = "",
  pages        = "361--397",
  month        = "",
  year         = "2004"
}

@inproceedings{AdamW2019,
  author       = "Ilya Loshchilov and Frank Hutter",
  title        = "Decoupled Weight Decay Regularization",
  booktitle    = "ICLR",
  publisher    = "",
  pages        = "",
  address      = "",
  month        = "",
  year         = "2019"
}

@article{MLL2014,
  author       = "Min{-}Ling Zhang and Zhi{-}Hua Zhou",
  title        = "A Review on Multi-Label Learning Algorithms",
  journal      = "{IEEE} Transactions on Knowledge and Data Engineering",
  volume       = "26",
  number       = "8",
  pages        = "1819--1837",
  month        = "",
  year         = "2014"
}

@inproceedings{MixMatch2019,
  author       = "David Berthelot and others",
  title        = "MixMatch: {A} Holistic Approach to Semi-Supervised Learning",
  booktitle    = "NeurIPS",
  pages        = "5050--5060",
  publisher    = "",
  address      = "",
  month        = "",
  year         = "2019"
}

@article{TCYB1,
  author       = {Zhiwen Yu and
                  Yidong Zhang and
                  Jane You and
                  C. L. Philip Chen and
                  Hau{-}San Wong and
                  Guoqiang Han and
                  Jun Zhang},
  title        = {Adaptive Semi-Supervised Classifier Ensemble for High Dimensional
                  Data Classification},
  journal      = {{IEEE} Transactions on Cybernetics},
  volume       = {49},
  number       = {2},
  pages        = {366--379},
  year         = {2019}
}

@article{TCYB2,
  author       = {Zhen Xu and
                  Ying Liu and
                  Chunguang Li},
  title        = {Distributed Semi-Supervised Learning With Missing Data},
  journal      = {{IEEE} Transactions on Cybernetics},
  volume       = {51},
  number       = {12},
  pages        = {6165--6178},
  year         = {2021}
}

@article{TCYB3,
  author       = {Jian Zhong and
                  Xiangping Zeng and
                  Wenming Cao and
                  Si Wu and
                  Cheng Liu and
                  Zhiwen Yu and
                  Hau{-}San Wong},
  title        = {Semisupervised Multiple Choice Learning for Ensemble Classification},
  journal      = {{IEEE} Transactions on Cybernetics},
  volume       = {52},
  number       = {5},
  pages        = {3658--3668},
  year         = {2022}
}

@article{TCYB4,
  author       = {Wandong Zhang and
                  Q. M. Jonathan Wu and
                  Yimin Yang},
  title        = {Semisupervised Manifold Regularization via a Subnetwork-Based Representation
                  Learning Model},
  journal      = {{IEEE} Transactions on Cybernetics},
  volume       = {53},
  number       = {11},
  pages        = {6923--6936},
  year         = {2023}
}

@article{TCYB5,
  author       = {Zhen Xu and
                  Ying Liu and
                  Chunguang Li},
  title        = {Distributed Information-Theoretic Semisupervised Learning for Multilabel
                  Classification},
  journal      = {{IEEE} Transactions on Cybernetics},
  volume       = {52},
  number       = {2},
  pages        = {821--835},
  year         = {2022}
}

@article{TCYB6,
  author       = {Lingling Zhang and
                  Minnan Luo and
                  Zhihui Li and
                  Feiping Nie and
                  Huaxiang Zhang and
                  Jun Liu and
                  Qinghua Zheng},
  title        = {Large-Scale Robust Semisupervised Classification},
  journal      = {{IEEE} Transactions on Cybernetics},
  volume       = {49},
  number       = {3},
  pages        = {907--917},
  year         = {2019}
}

@article{TCYB7,
  author       = {Yuan Xie and
                  Wensheng Zhang and
                  Yanyun Qu and
                  Longquan Dai and
                  Dacheng Tao},
  title        = {Hyper-Laplacian Regularized Multilinear Multiview Self-Representations
                  for Clustering and Semisupervised Learning},
  journal      = {{IEEE} Transactions on Cybernetics},
  volume       = {50},
  number       = {2},
  pages        = {572--586},
  year         = {2020}
}

@article{TCYB8,
  author       = {Jingchen Ke and
                  Chen Gong and
                  Tongliang Liu and
                  Lin Zhao and
                  Jian Yang and
                  Dacheng Tao},
  title        = {Laplacian Welsch Regularization for Robust Semisupervised Learning},
  journal      = {{IEEE} Transactions on Cybernetics},
  volume       = {52},
  number       = {1},
  pages        = {164--177},
  year         = {2022}
}

@article{TCYB9,
  author       = {Jian Chen and
                  Lan Du and
                  Leiyao Liao},
  title        = {Discriminative Mixture Variational Autoencoder for Semisupervised
                  Classification},
  journal      = {{IEEE} Transactions on Cybernetics},
  volume       = {52},
  number       = {5},
  pages        = {3032--3046},
  year         = {2022}
}

@inproceedings{AM1,
  author       = {Ting{-}En Lin and
                  Hua Xu},
  title        = {Deep Unknown Intent Detection with Margin Loss},
  booktitle    = {ACL},
  pages        = {5491--5496},
  publisher    = "",
  address      = "",
  month        = "",
  year         = {2019}
}

@inproceedings{AM2,
  author       = {Yinfei Yang and
                  Gustavo Hern{\'{a}}ndez {\'{A}}brego and
                  Steve Yuan and
                  Mandy Guo and
                  Qinlan Shen and
                  Daniel Cer and
                  Yun{-}Hsuan Sung and
                  Brian Strope and
                  Ray Kurzweil},
  title        = {Improving Multilingual Sentence Embedding using Bi-directional Dual
                  Encoder with Additive Margin Softmax},
  booktitle    = {IJCAI},
  pages        = {5370--5378},
  publisher    = {},
  address      = "",
  month        = "",
  year         = {2019}
}
